\documentclass[journal]{IEEEtran}
\usepackage{cite}
\usepackage{xspace}
\usepackage{graphicx}
\usepackage{amsmath}
\usepackage{color}
\usepackage{tcolorbox}
\usepackage{pgf}
\usepackage{tikz}
\usepackage{verbatim}
\usepackage[utf8]{inputenc}
\usetikzlibrary{arrows,automata}
\usetikzlibrary{positioning}

\tikzset{
    state/.style={
           rectangle,
           rounded corners,
           draw=black, very thick,
           minimum height=2em,
           inner sep=2pt,
           text centered,
           },
}

\ifCLASSINFOpdf
\else
\fi

\DeclareMathOperator*{\otherwise}{otherwise}
\DeclareMathOperator*{\ifff}{if}
\DeclareMathOperator*{\possibility}{probability}
\DeclareMathOperator*{\with}{with}
\DeclareMathOperator*{\clip}{clip}
\DeclareMathOperator*{\argmin}{arg \, min}
\DeclareMathOperator*{\argmax}{arg \, max}

\newcommand{\etal}{\textit{et al.}\xspace}
\newcommand{\nsp}{\textit{non-structure pruning}\xspace}
\newcommand{\spt}{\textit{structure pruning}\xspace}

\begin{document}
\title{Bringing AI To Edge: From Deep Learning's Perspective}

\author{Di Liu, Hao Kong, Xiangzhong Luo, Weichen Liu, Ravi Subramaniam
                   \thanks{
                   Di Liu, Hao Kong, and Xiangzhong Luo contribute equally to this work.
                   
                   This research was conducted in collaboration with HP Inc. and supported by National Research Foundation (NRF) Singapore and the Singapore Government through the Industry Alignment Fund-Industry Collaboration Projects Grant (I1801E0028). This work is also partially supported by NTU NAP M4082282 and SUG M4082087, Singapore.
                   
                  Di Liu is with HP-NTU Digital Manufacturing Corporate Lab, Nanyang Technological University, Singapore. (Email: liu.di@ntu.edu.sg)
                  
                  Hao Kong and Weichen Liu are with HP-NTU Digital Manufacturing Corporate Lab, Nanyang Technological University, Singapore, and School of Computer Science and Engineering, Nanyang Technological University, Singapore. (Email: kong.hao@ntu.edu.sg, liu@ntu.edu.sg)
                  
                  Xiangzhong Luo is with School of Computer Science and Engineering, Nanyang Technological University, Singapore. (Email: xiangzho001@e.ntu.edu.sg)
                  
                  Ravi Subramaniam is with Innovations and Experiences–Business Personal Systems, HP Inc., USA. (Email: ravi.subramaniam@hp.com)
                  }

}

\maketitle

\begin{abstract}
% Edge computing and deep learning are the two most prevailing research fields.
% Edge computing is a promising extension of cloud computing, providing low latency and privacy-protected services for users. On the other hand, deep learning techniques are advancing many applications, like voice assistant, machine translation, autonomous driving, etc, giving us a more \textit{intelligent} life. 
Edge computing and artificial intelligence (AI), especially deep learning for nowadays, are gradually intersecting to build a novel system, called \textit{edge intelligence}. 
However, 
the development of edge intelligence systems encounters some challenges, 
and one of these challenges is the \textit{computational gap} between computation-intensive deep learning algorithms and less-capable edge systems.
Due to the computational gap, many edge intelligence systems cannot meet the expected performance requirements. To bridge the gap, 
a plethora of deep learning techniques and optimization methods are proposed in the past years: light-weight deep learning models, network compression, and efficient neural architecture search. Although some reviews or surveys have partially covered this large body of literature,   
we lack a systematic and comprehensive review to discuss all aspects of these deep learning techniques which are critical for edge intelligence implementation. As various and diverse methods which are applicable to edge systems are proposed intensively, a holistic review would enable edge computing engineers and community to know the state-of-the-art deep learning techniques which are instrumental for edge intelligence and to facilitate the development of edge intelligence systems.
This paper surveys the representative and latest deep learning techniques that are useful for edge intelligence systems, including hand-crafted models, model compression, hardware-aware neural architecture search and adaptive deep learning models.  
Finally, based on observations and simple experiments we conducted, we discuss some future directions.

\end{abstract}


\section{Introduction}

In 2012, AlexNet \cite{Alexnet} broke the record of ImageNet LSVRC contest \cite{imagenet}, improving the prediction accuracy by a large margin of 10\%, which marked the milestone for the booming development of deep learning (DL) techniques, or more specifically deep neural network (DNN) \cite{Goodfellow2016}. 
In the subsequent years, as the increasing number of DNN applications\cite{lecun2015deep} emerges, designing DNN-based systems attracts a lot of attention and efforts from academia and industry, spanning from advanced image manipulation and enhancement and convenient voice assistant on mobile phones to Level-4 autonomous driving systems on automotives  \cite{bojarski2016end}. 
AlphaGo \cite{silver2016mastering} is another prominent example which defeated the top professional player in the ancient Chinese game GO, deemed impossible before due to its extremely high complexity.

The successful application of DNN models relies on two stages: \textbf{training} and \textbf{inference/deployment}. 
Training indicates the procedure of learning a predictive model from a huge amount of labeled data, while inference denotes the procedure of using the trained model upon new data to make an accurate prediction. 
The training procedure involves a complex learning process requiring powerful computational units and a huge amount of data and spending considerable time. For instance, training ResNet50 \cite{he2016deep} with ImageNet on 8 Nvidia Tesla P100 takes 29 hours \cite{goyal2017accurate}, where ImageNet dataset \cite{imagenet} has ~1.28 M images of 1000 classes for training. Recently, researchers also measure the environmental effect of the training procedure and indicate that training DNN consumes excessive energy and emits substantial CO$_2$ \cite{strubell2019energy} and environment-sustainable DNN models should be considered. 

The trained DNN model is exploited to make prediction on new data, and this procedure is usually called DNN \textit{inference}.  
Many companies implement their DNN inference on servers to provide convenient services for their customers such as voice-assistant, machine translation, image retrieval, etc \cite{siri}. 
However, inference from cloud suffers from \textit{responsiveness} (e.g., latency) and \textit{privacy} issues which are critical in some scenarios. 
For instance, DNNs are seen to be a pivotal technique for self-driving cars, in which most of executions are subject to highly rigorous real-time constraints \cite{bojarski2016end,baruah2020achieving}, but the high and uncertain communication overhead makes it difficult to satisfy the temporal requirement. 
In addition, some inference is conducted on confidential data, e.g., manufacturing and production data, and uploading these data to cloud may risk data leakage which will severely hurt their business. 
To address these concerns, \textit{edge computing} is proposed \cite{shi2016edge}. 

\begin{figure}
    \centering
    \includegraphics[width=\columnwidth]{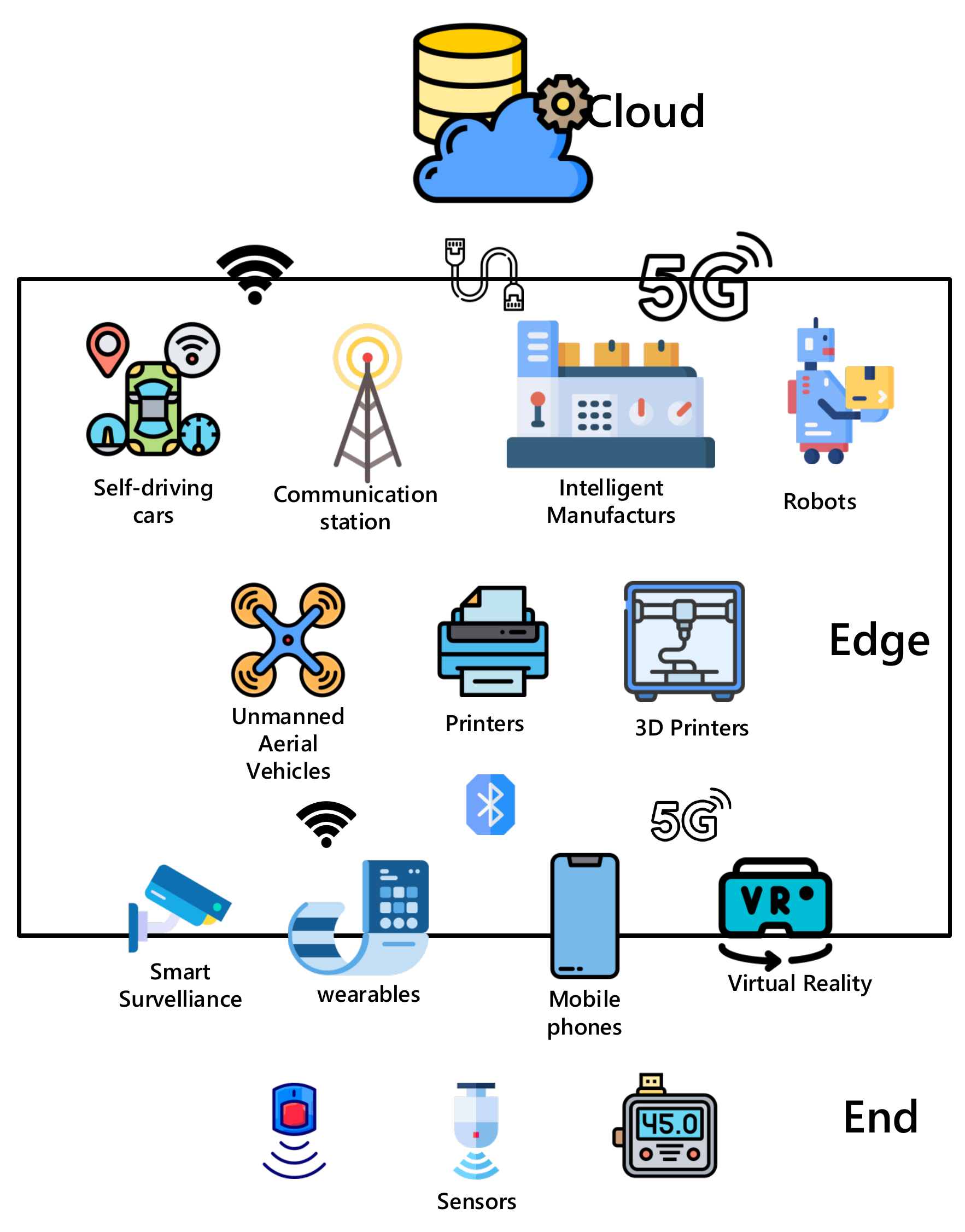}
    \caption{The wide spectrum of Edge Computing}
    \label{fig:overview}
\end{figure}
% In addition, to pursue higher accuracy, deep learning model are 
 
Edge computing systems are placed at the proximity of data producer, like sensors, end-users, etc,  to rapidly and locally process data. 
The emergence of edge computing paves the way for pervasive intelligent systems \cite{edgeintelligenc}.
IEEE Computer Society identifies  `Artificial Intelligence at the edge' as one of the top 12 technology trends to reach adoption in 2020 \cite{ieeetrend}. 
Edge systems have a broad spectrum of hardware features, from powerful computation units in communication stations and self-driving cars to small and battery-supplied smart phones and wearable devices. Fig. \ref{fig:overview} plots the overview of edge computing.
In different scenarios, edge systems are subject to diverse performance and physical constraints, e.g., latency and power constraints for self-driving cars, computational capability and power constraints for UAVs, etc. However, DNN models, the current core component of artificial intelligence, are highly computation-intensive and are even growing `\textit{wider}' (more filters within a layer) and `\textit{deeper}' (more layers) with billions of parameters and millions of float point operations (FLOPs) (Some preliminaries about DNN are given in Section \ref{section:prem}). 
Such `\textit{deeper}' models provide high accuracy at the cost of highly computational complexity. 
This unavoidably leads to \textit{`computational gap'} between DNN models and less-capable edge systems. 
 
Many approaches are proposed to bridge the `\textit{computational gap}' between the complex models and resource-constrained edge systems. From hardware perspective, different accelerators are proposed and many companies are dedicated to develop specific hardware architectures to speed up the execution of DNN models, such as application-specific integrated circuits (ASICs) Tensor Processing Unit (TPU) \cite{tpu} and DianNao family \cite{diannao}, and FPGA-based accelerators ESE \cite{han2017ese}\cite{fpgasurvey}. 
However, developing chips is a costly process, 30-80 million dollars and 2-3 years to develop \cite{feldman2019era}. Besides, the benefit of porting DNN applications to these new computing units is still under doubt. The research from Facebook shows the performance gain by porting DNN applications to specific hardware accelerators may not be able to compensate the significant engineering effort spent due to the poor ecosystems and program-ability \cite{facebook}. In addition, some customized edge accelerators achieve high efficiency at the cost of generality, e.g., edge TPU only supports a limited 
number of operators \cite{edgetpu} and some new DNN architectures cannot be well-supported. 

On the other hand, software is known to be flexible and less costly to develop in comparison with hardware. For efficient edge intelligence systems, from software perspective, the goal is to design proper DNN models which can be fit on edge systems while guaranteeing the required performance and maximizing accuracy. 
Different methodologies are devised to approach this goal, like novel light-weight DNN model design, model compression, and the emergent neural architecture search (NAS) \cite{Zoph2016nas}, all of which facilitate the development and application of DNN models on edge systems.
% The obvious advantage from software perspective, i.e., deep learning techniques, is that since DNN research and design are evolving very fast, it can easily modify DNN structure for different hardware platforms under different objectives, thereby reducing the time-to-market and engineering cost. 
In the past decade, many DNN breakthroughs have significantly improved DNNs' accuracy and performance and we believe the innovation from deep learning techniques will still be the key ingredient for the emergent edge intelligence.

% Intelligent edge systems are poised to boost from 2020 onward, since the inference part of DL will gradually dominate the market.

\subsection{Motivation Behind This Paper}

There exist several excellent reviews about edge computing and edge intelligence. Zhou \etal \cite{edgeintelligenc} discuss the great potential of edge intelligence and point out some future research directions. Wang \etal \cite{wang2020convergence} mainly review the edge intelligence systems from communication systems' perspective. Chen and Ran in \cite{deepedge} discuss the diverse applications when deep learning techniques are integrated into edge computing. All of these reviews only have a small fraction to discuss the edge DNN design which is the key element of edge intelligence systems. 

The vast DNN design methods applicable to edge systems can be broadly classified into three categories: \textit{hand-crafted models}, \textit{model compression} and \textit{hardware-aware NAS}. 
Some prior DNN reviews cover one of the three categories, like \cite{cheng2018model}\cite{deng2020model} for model compression, \cite{sze2017efficient} for early works on efficient DNN processing and \cite{elsken2019neural} for NAS\footnote{This survey is for the general NAS, not the hardware-aware NAS.}
but there is no review discussing all three categories which in future will be integrated seamlessly to design effective and efficient DNNs in the new edge intelligence era. 
Considering the anticipated emergence of numerous edge intelligence systems in the next years \cite{edgeintelligenc}, this motivates us to comprehensively review the relevant works about edge DNN design in a holistic fashion. 

In addition, except the static DNN design for edge systems, we also discuss adaptive DNN models which are also not reviewed systematically in previous articles. 
Edge systems installed in real world operate in highly dynamic environment, while it still needs to guarantee a certain degree of quality of service (QoS) and performance, like real-time requirement. However, static DNN models are unable to achieve such guarantee under dynamic environment \cite{facebook}. Hence, we need models which are able to adapt their computation to achieve a good trade-off between accuracy and efficiency. 

\subsection{Structure of This Paper}
In general, we consider this paper as a good complementary for other related surveys or reviews in the field of edge computing and hope this paper facilitates a profound understanding from deep learning's perspective, which plays a pivotal role in implementing edge intelligence systems. 
As shown in \cite{dean2018new}, around 50 papers were uploaded to Arxiv per day in 2018, discussing DNN related topics, i.e., in total more than 15000 papers per year. It is supposed to have more now. AlexNet which broke the record in 2012 now is rarely deemed as an important reference approach for experimental comparison. 
Therefore, it is impossible to review all related papers.  In this paper, due to the space limitation, we strive to cover the most relevant, representative, and latest works\footnote{We include peer-reviewed papers until 2020} which, we believe, are adequate for readers to fully understand the development, motivation, and latest trend of these DL techniques applicable for edge systems and to have a full view of these techniques. 
% Deep learning and edge computing are both relative new concepts, while deep learning techniques start to bloom since 2012 \cite{Alexnet} and the concept of edge computing is proposed around 2015 \cite{shi2016edge}.  
% DNN research develops rapidly, so we review the relevant works until 2020. 
Finally, we provide some of our thoughts about designing DNNs for edge devices based on some observations and experiments we have conducted. 

\textbf{Scope of this paper}: Edge computing is an emergent research topic, consisting of many interesting and challenging research problems, like edge caching \cite{liu2016caching} and computation offloading \cite{mach2017mobile,yu2020joint}. Prior review or surveys have profoundly discussed these topics. In this paper, we only focus on deep learning techniques which aid in developing edge intelligence systems. For other related topics, we refer interesting readers to other excellent literature \cite{mach2017mobile,liu2016caching,wang2020convergence}.

The reminder of this paper is organized as follows: 
\begin{itemize}
    \item Section \ref{section:prem} introduces the preliminaries of DNN models to facilitate the understanding of techniques presented in subsequent sections.
    \item Section \ref{section:hand} discusses works designing novel DNN architectures for light-weight DNN models.
    \item  Section \ref{section:compress} reviews network compression methods, including network pruning, quantization, and knowledge distillation.
    \item Section \ref{section:nas} discusses NAS techniques aiming to design and customize efficient DNN models for resource-constrained systems, like some edge systems. 
    \item Section \ref{sec:adaptive} discusses adaptive DNN models which may adapt the model computation under dynamic environment.
    \item Section \ref{section:discussion} demonstrates our thoughts on future edge DNN designs based on our observation and experimental results. 
    \item Section \ref{section:conclusion} concludes this paper.
\end{itemize} 
   
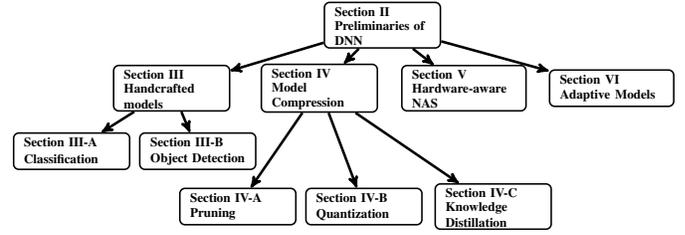
\begin{figure}
    \centering
    \resizebox{\columnwidth}{!}{\begin{tikzpicture}[->,>=stealth']

 % State: ACK with different content
 \node[state,    	% layout (defined above)
  text width=3cm, 	% max text width
  yshift=2cm, 
  xshift=2.5cm,% move 2cm in y
%   right of=HC, 	% Position is to the right of HC
  node distance=6.5cm, 	% distance to HC
  anchor=center] (PM) 	% posistion relative to the center of the 'box'
 {%
 \begin{tabular}{l}
      \textbf{Section \ref{section:prem}}\\
      \parbox{2.8cm}{\textbf{Preliminaries of DNN} }
 \end{tabular}
 };

 % Position of HC 
 % Use previously defined 'state' as layout (see above)
 % use tabular for content to get columns/rows
 % parbox to limit width of the listing
 \node[state,
 text width=3cm,
 below left of=PM,
 xshift=-5cm,
 yshift=-1cm
 ] (HC) 
 {\begin{tabular}{l}
 \textbf{Section \ref{section:hand}}\\
  \parbox{2.8cm}{\textbf{Handcrafted models}}
 \end{tabular}
 };
 
 % STATE prem
 \node[state,
  below left of=PM,
  yshift=-1cm,
  xshift=-1cm,
  anchor=center,
  text width=3cm] (MC) 
 {%
 \begin{tabular}{l}
  \textbf{Section \ref{section:compress}}\\
  \parbox{2.8cm}{\textbf{Model Compression}}
 \end{tabular}
 };

 % STATE EPC
 \node[state,
  below right of=PM,
  yshift=-1cm,
  xshift=1.5cm,
%   node distance=5cm,
  anchor=center] (NAS) 
 {%
 \begin{tabular}{l}
  \textbf{Section \ref{section:nas}}\\
  \parbox{2.8cm}{\textbf{Hardware-aware NAS}}
 \end{tabular}
 };

  \node[state,
  below right of=PM,
  yshift=-1cm,
  xshift=5.5cm,
%   node distance=5cm,
  anchor=center] (AM) 
 {%
 \begin{tabular}{l}
  \textbf{Section \ref{sec:adaptive}}\\
  \parbox{2.8cm}{\textbf{Adaptive Models}}
 \end{tabular}
 };
 
   \node[state,
 text width=3cm,
 below left of=HC,
 yshift=-1cm,
 xshift=-2cm
 ] (CC) 
 {\begin{tabular}{l}
    \textbf{Section \ref{hand:class}}\\
    \parbox{2.8cm}{\textbf{Classification}}
 \end{tabular}
 }; 
 
  \node[state,
 text width=3cm,
yshift=-1cm,
 below right of=HC
 ] (OD) 
 {\begin{tabular}{l}
    \textbf{Section \ref{hand:od}}\\
    \textbf{Object Detection}
 \end{tabular}
 }; 
 
   \node[state,
 text width=3cm,
xshift=-1.5cm,
yshift=-2.5cm,
 below left of=MC
 ] (PU) 
 {\begin{tabular}{l}
    \textbf{Section \ref{compress:pruning}}\\
    \parbox{2.8cm}{\textbf{Pruning}}
 \end{tabular}
 }; 
 
    \node[state,
 text width=3cm,
xshift=0.5cm,
yshift=-2.5cm,
 below right of=MC
 ] (QU) 
 {\begin{tabular}{l}
    \textbf{Section \ref{compress:quant}}\\
    \parbox{2.8cm}{\textbf{Quantization}}
 \end{tabular}
 }; 
 
 \node[state,
 text width=3cm,
xshift=4cm,
yshift=-2.5cm,
 below right of=MC
 ] (KD)
 {\begin{tabular}{l}
    \textbf{Section \ref{compress:kd}}\\
    \parbox{2.8cm}{\textbf{Knowledge Distillation}}
 \end{tabular}
 };

 % draw the paths and and print some Text below/above the graph
 \path (PM)[line width=2pt] 	edge   (HC)
 (PM)     	edge  (MC)
 (PM)     	edge  (NAS)
 (PM)     	edge  (AM)
 (HC)     	edge  (CC)
 (HC)     	edge  (OD)
 (MC)     	edge  (PU)
 (MC)     	edge  (QU)
 (MC)     	edge  (KD)
 ;

\end{tikzpicture}}
    \caption{The schematic structure of this paper}
    \label{fig:content}
\end{figure}

Fig \ref{fig:content} shows the detailed structure of this paper.

% Neural architecture search (NAS) has become an interesting topic in DL model design. Instead of using experts' knowledge and insight to design DL models, NAS strive to automatically find an optimal deep neural architecture for a specific objective. 
\section{Preliminaries}
\label{section:prem}
\begin{figure*}[!th]
\centering
\includegraphics[width=\textwidth]{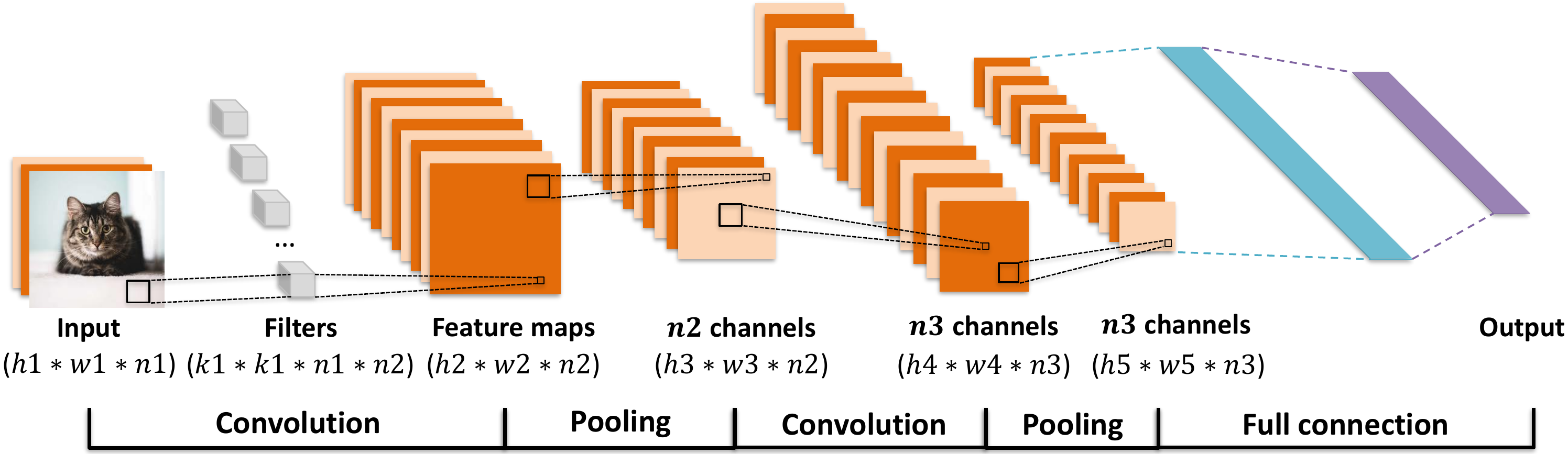}
\caption{The overview of a DNN/CNN model}
\label{figure:dnn}
\end{figure*}

In this section, we introduce some basic knowledge of DNNs for better understanding the techniques and approaches discussed in subsequent sections.
% DNNs are the most commonly used artificial neural networks in AI applications nowadays, which are usually composed of multiple/many layers of operations. 
In this paper, we mainly review the models and techniques proposed for computer vision applications and there are two reasons for this. First,  currently computer vision models are the major application for edge intelligence systems; Second, although some recent studies start to investigate how to design lightweight natural language processing (NLP) models for edge devices like, \cite{litetransformer}\cite{tinybert}\cite{mobilebert}, they are far from mature like models and techniques for computer vision tasks. Hence, in this paper, we use DNN models of computer vision as the major example for our presentation.
 
For computer vision applications, when referring to DNNs, we mean convolutional neural networks (CNN). Throughout this paper, we may use DNN and CNN interchangeably. 
CNNs usually have many convolutional layers, pooling layers, activation layers, and a couple of fully connected layers \cite{Goodfellow2016}. 
Figure \ref{figure:dnn} shows a simple CNN with two convolutional layers and one fully connected layer. 
The convolutional layer is the core ingredient in CNNs, which extracts the patterns/features from input data at different granularity. Convolutional layers account for the major resource and time cost of a CNN model. To better illustrate how the convolutional layer works, we give some terminologies about convolutional layers as follows:
\begin{itemize}
    \item \textbf{Kernel} -- A kernel is a 2D square matrix with size like $3\times3$, $5\times5$, $7\times7$ and a convolutional operand.
    \item \textbf{Filter} -- A filter is a collection of kernels with size of $k\times k \times c$, where $k$ is the kernel size and $c$ is equal to the number of input channels. A filter is convolved with all input channels to derive a new channel/feature map/activation map.  
    % For single-channel layer, the filter is the same with the kernel. For multi-channel layers, the filter is a collection of kernels. Kernels inside the filter are applied onto corresponding channels of the input to generate an output feature map.
    \item \textbf{Feature map/channel/activation} -- Feature map, channel, and activation have the same meaning in CNNs. 
    % \footnote{Now, many hand-crafted models are designed by stacking the same block, where a block contains a couple of operations.} 
    A feature map is a 2D feature matrix generated by a filter. 
    % Feature maps are the output of the previous layer, and at the same time, they are also the input of the next layer.
    % \item \textbf{Channel} -- Actually, each feature map is also called a channel. For multi-channel layers, every channel contains different information. For example, an RGB image has three channels and each channel stores the information of one primary color.

\end{itemize}
As demonstrated in Figure \ref{figure:dnn}, the input of the first convolutional layer is an image with $n_1$ channels, and each channel is a 2D array with height $h_1$ and width $w_1$. Correspondingly, there are $n_1$ kernels in each filter, and the kernel size is $k_1 \times k_1$. There are $n_2$ filters, and thus $n_2$ feature maps will be generated after the first convolution operation and the size of the first feature maps is $(h_2\times w_2\times n_2)$, where $h_2$ and $w_2$ are the height and weight of the first feature maps, respectively. If the padding is used during convolution, then $h_1=h_2$ and $w_1=w_2$. Otherwise, $h_1>h_2$ and $w_1 > w_2$. For more details, interesting readers are referred to \cite{Goodfellow2016}.

% At each convolutional layer, there are two groups of feature maps containing the information of the original input, where each feature map acts as a channel for data transmission. For example, the input of the DNN demonstrated in Figure \ref{figure:dnn} has $n1$ channels.
% There are a set of filters being employed to transform information between feature maps at different hierarchies. The first two dimensions of the shape of filters is the height and width of the kernels in filters, while the third and forth shape parameters indicate the number of kernels in each filter and the number of filters, respectively.
% On the one hand, each filter integrates all channels of the input feature maps into one output feature map, enabling information communication among different channels. On the other hand, different filters inside the convolutional layer create new channels of the output feature maps, and further control the number of feature dimensions.

Usually, a convolutional layer is followed by a pool layer and an activation layer. In Fig. \ref{figure:dnn} , we omit activation layers. Activation functions are used to introduce non-linearity into neural network, and the common activation functions are rectify linear units (ReLu)\cite{relu}, sigmoid, etc. 
The pooling layer down-samples the feature map to reduce the spatial size of the feature map and increase its receptive field. The max pooling \cite{maxpooling} and the average pooling are the two common pooling methods.
A DNN model usually ends up with a couple of fully connected layers,
which is used to fuse the feature information from the last convolutional layer and predict the classification of the input image. 
To improve the performance and training speed, modern DNN models also have other operation layers, like batch normalization layer \cite{bn}, squeeze-and-excitation layer \cite{squeeze-excitation}, etc. 
\section{Lightweight Network Design}
\label{section:hand}
AlexNet \cite{Alexnet} marks the milestone of rapid development of DNN models while researchers find that the bigger a DNN model is, the better accuracy the model can provide. As a result, this incentivizes the emergence of increasingly complicated DNN models \cite{computational_limit}. However, the highly computational complexity of DNN models hinders them from being efficiently deployed on resource-constraint devices, e.g., edge or IoT devices. 
A promising way to solve this problem is to design novel neural architectures/operators which are more efficient on edge systems while not compromising accuracy. 
% Therefore, researchers propose some new structures and computation-efficient operations to devise light-weight models for edge systems. 
In this section, we review the works which target to manually design lightweight models. For these hand-craft models, we classify them into two categories based on their applications: \textit{classification} and \textit{object-detection}. These two applications are the major deployment scenarios for edge intelligence systems. 
% Although some light-weight models for nature language processing are also desired, the majority of efforts for improving the efficiency of DNN models are placed on these two applications. 

\subsection{Classification}
\label{hand:class}
Classification is the fundamental and most critical task in computer vision and it is also the first application in which DNNs demonstrated its huge potential \cite{Alexnet}. 
Moreover, classification models play a core role in other computer vision tasks, such as semantic segmentation and object detection, where classification models serve as the backbone to extract features from images and provide essential feature information for segmentation and detection.

SqueezeNet \cite{iandola2016squeezenet} is one of the early works towards designing DNNs for resource-constrained hardware. The core idea in SqueezeNet is a new computational module, \textit{Fire module}, in which convolutional operations are split into the squeeze layer and expand layer and some $3\times 3$ convolutional operations are replace by low-complexity $1\times 1$ convolution operations. SqueezeNet achieves AlexNet-level accuracy while greatly reducing the model complexity. The applicability of SqueezeNet is validated on real FPGA with small on-chip memory. Gschwend \etal \cite{gschwend2016zynqnet} develop a variant of SqueezeNet and implement it on an FPGA. It shows the SqueezeNet-like model can be entirely fit within the FPGA on-chip memory. As the result, it significantly eliminates the off-chip memory access overhead and in turn improves the inference performance.

MobileNet series \cite{howard2017mobilenets,sandler2018mobilenetv2,howard2019searching} are another prominent DNN models targeting resource-limited devices. MobileNet \cite{howard2017mobilenets} replaces the conventional convolutional operation with more efficient depth-wise separable convolution operation proposed in \cite{chollet2017xception} to reduce the computational cost. Depth-wise separable convolution operation factorizes a conventional $k \times k \times n$ convolution into a $k\times k \times 1$ depth-wise convolution and a $1 \times 1 \times n$ point-wise convolution. Each input channel is convolved with the depth-wise convolution operator and the point-wise convolution linearly combines all results from depth-wise convolution to generate one channel/feature map. Depth-wise separable convolution can significantly reduce the computational complexity, thus shortening inference time on edge devices. 
On the top of MobileNet, MobileNetV2 \cite{sandler2018mobilenetv2} adds linear bottleneck and inverted residual block to improve both accuracy and performance. The latest MobileNetV3 \cite{howard2019searching} combines NAS and NetAdapt \cite{yang2018netadapt} to design a more accurate and efficient network architecture. 

\begin{table}[h]
    \centering
    \caption{Comparison of hand-crafted models. Each model has a scaling factor to scale up the number of channels. Here we only report the results of the scaling factor=1.}
    \begin{tabular}{c|c|c|c|c}
    \hline
        Model & Year & Parameters & FLOPs/MACs & Accuracy \\ \hline
         SqueezeNet \cite{iandola2016squeezenet} & 2016 & 1.24M & & 60.4\% \\
         Mobilenet \cite{howard2017mobilenets} & 2017 & 4.2M & /569M & 70.6\% \\
         MobilenetV2 \cite{sandler2018mobilenetv2} & 2018 & 3.4M & /300M &  72\% \\
         MobilenetV2-large \cite{sandler2018mobilenetv2} & 2018 & 6.9M & /585M &  74.7\% \\
         MobilenetV3-small \cite{howard2019searching} & 2019 & 2.5M & /56M &  67.4\% \\
         MobilenetV3-large \cite{howard2019searching} & 2019 & 5.4M & /300M &  75.2\% \\
         ShuffleNet \cite{zhang2018shufflenet}& 2018 & 3.4M & /292M &  71.5\% \\
         ShuffleNetV2 \cite{ma2018shufflenet} & 2018 & 2.3M & 146M/ & 69.4\%\\
         EfficientNet-B0 \cite{tan2019efficient} & 2019 & 5.3M & 390M/ & 77.1\% \\
         GhostNet \cite{han2019ghostnet} & 2020 & 5.2M & 141M/ &  73.9\% \\
         \hline
    \end{tabular}
    \label{tab:hc_comparison}
\end{table}
% Zhang \etal in \cite{zhang2018shufflenet} observe \textit{group convolution} can be used to reduce computational cost, but the information propagation method in traditional \textit{group convolution} method prevents each group from obtaining information from other groups and it in turn hurts accuracy of these models. Thus, 

% Other work also deploy the same concept of replacing the complex operators with more efficient operation. 
% Other light-weight models propose different ways to reduce their model complexity. 
ShuffleNet \cite{zhang2018shufflenet} deploys \textit{group convolution} and \textit{channel shuffle} to reduce the computational complexity while retaining high accuracy. 
ShuffleNetV2 \cite{ma2018shufflenet} empirically observes four principles for designing efficient DNNs and proposes \textit{channel split} to improve accuracy and performance.  
GhostNet \cite{han2019ghostnet} proposes a \textit{ghost module} based on an observation that some features in convolutional layers are highly correlated. Thus, it first uses standard convolutional operation to obtain a few intrinsic features and then generates more features from the intrinsic features with cheap linear operations.

EfficientNet \cite{tan2019efficient} investigates the influence of three scaling dimensions in DNN models, i.e., depth/layer scaling, channel scaling and resolution scaling, and proposes a compound scaling method such that given a DNN model and a target computational complexity (i.e., FLOPs) it can effectively adjust the three dimensions of a model to improve its predictive performance. 
The scaled models achieves a comparable or better accuracy while having fewer parameters. 
% EfficientNet does not aim at resource constrained systems, but the dimension scaling is believed to play an important role in managing model complexity. 

\noindent\textbf{Discussion:} Table \ref{tab:hc_comparison} summarizes the models discussed in this section. Significant progresses have been made in lightweight DNN models for classification task, and classification DNN models are also the key component for other CV tasks, like segmentation, detection, etc. 
The manual design of lightweight DNN models is considerably dependent on experts' knowledge and also needs a time-consuming hyperparameter exploration. 
As \textit{automated machine learning} (AutoML\footnote{https://https://www.automl.org/}) \cite{automl} techniques, like NAS, hyperparameter optimization, emerge to aid in designing DNN models, designers can put their focus on designing effective and efficient DNN modules or operations. Then the new operations can be used as the fundamental elements to generate new models for edge systems. 
% is also a well-studied topic in DNN
% s impossible to enumerate all works in this paper. We only select some representative and widely-used models like MobileNet and ShuffleNet and some latest models, EfficientNet and GhostNet. 
% Since convolutional layers account for the majority of execution time \cite{he2015convolutional}, all of these works share a common goal, i.e., designing novel and light-weight methods to replace the conventional convolutional layers. Novel neural architectures or layer structures are the foundation for constructing edge DNN models, 
For instance, the latest MobileNetV3 exploits NAS with the novel architectures from MobileNetV2 and MnasNet \cite{tan2019mnasnet} to find accurate and efficient DNN model for mobile setting. 
We think this is becoming a mainstream trend to adopt NAS with novel lightweight operators to design edge DNN model. 
However, existing methods overlook the impact of deployed hardware, on which some operators cannot be supported or effectively executed. 
\textit{Thus, for the future DNN classification models, especially for  edge systems, it is important to design models in a hardware-aware fashion. }

% These hand-crafted and light-weighted models actually serves as foundation for other mhave proven their effectiveness and efficiency in many applications. 
% As new research indicate, NAS may gradually replace 

\begin{table*}[!h]
    \centering
    \caption{Hand-crafted object detection models summary. All information of this table is from the original paper or the paper which compare these models. Experimental results are upon two widely-used benchmark, PASCAL VOC2007\cite{pascal} and MS COCO\cite{mscoco}.}
    \begin{tabular}{c|c|c|c|c|c|c|c|c}
    \hline
        Model & Year & Methods & Input & Parameters & FLOPs/MACs & mAP(PASCAL) & mAP(COCO) & Tested Platform\\ \hline
        SqueezeDet \cite{squeezedet} & 2017 & One-stage &$1242\times375$& - & 9.7B/& - & - & GPU \\
        Tiny-YOLO \cite{yolov3}& 2018 & One-stage &  $416\times416$ &15.12M & 3.49B/ & 57.1\% & -  & GPU  \\
        Tiny-DSOD \cite{li2018tiny}& 2018 & One-stage &  $300\times300$& 1.15M & 1.12B/ & 72.1\%& 23.2\%& GPU  \\
        Tiny-SSD \cite{tinyssd} & 2018 & One-stage&  $300\times300$ & 1.13M&/571.09M & 61.3\%& - & GPU  \\
        Pelee \cite{wang2018pelee} & 2018 & One-stage&$320\times320$  &5.98M &1.21B/& 70.9\%& 22.4\%& GPU/Edge GPU \\
       Light-Head RCNN \cite{lightheadrcnn} & 2017 & Two-stage &$800\times 1200$ & - &5.65BM/&75.1\% & - & GPU \\
       ThunderNet \cite{Qin_2019_ICCV} & 2019& Two-stage &$320\times320$ & - &4.61M/&75.1\% & - & GPU/Edge  CPU \\
       EfficientDet \cite{tan2020efficientdet} & 2020 & Two-stage & - & 8.1M* & 11B*/ & - &43.0\%* & GPU \\
       
       \hline
    \end{tabular}
*EfficientDet has models with different complexity. We report the results of EfficientDet D2 which may be applicable to edge systems. 
    \label{tab:my_label}
\end{table*}

\subsection{Object Detection}
\label{hand:od}
Object detection is another vital field in DNN research. 
Besides predicting the category of an input image, object detection locates objects in the input image by drawing bounding boxes. 
Generally, we can classify object detection methods into two categories: one-stage method and two-stage method. One-stage methods predict object classification and localization in one single forward pass. YOLO \cite{yolo} and SSD \cite{ssd} are two widely-used examples of one-stage methods; Two-stage methods first deploy a backbone CNN network (usually for classification) to extract features from input images and then a detection part uses the features extracted from the backbone network to localize objects. RCNN \cite{r-cnn}, Fast RCNN \cite{fastrcnn}, and Faster RCNN \cite{fasterrcnn} are representative examples of two-stage methods. 
  
To boost the efficiency of object detection on resource-limited devices, some works strive to reduce the computational complexity of backbone part of object detection models, like Tiny-dsod \cite{li2018tiny} for  DSOD \cite{dsod}, Tiny-SSD \cite{tinyssd} for  SSD \cite{ssd}, and Tiny-YOLO \cite{yolov3} for YOLO. Other efforts combines SSD framework with lightweight CNNs, like MobileNet, ShuffleNet,  SequeezeNet, etc to improve efficiency.  
SequeezeDet \cite{squeezedet} implements an object detection model by using SequeezeNet \cite{iandola2016squeezenet} as the backbone to improve the efficiency while not significantly compromising accuracy. 
Pelee \cite{wang2018pelee}  combines an improved SSD framework with its optimized PeleeNet to improve efficiency of object detection. Since one-stage methods are less computational than two-stage methods, all of these methods mentioned above target one-stage method for improving efficiency of object detection DNNs. 

Few efforts strive to improve efficiency of two-stage methods. Li \etal \cite{lightheadrcnn} replace the heavy-head in RCNN framework to speed up execution. Qin \etal \cite{Qin_2019_ICCV} observe a good configuration of input resolution, backbone network, and detection head can readuce the complexity of two-stage methods while maintaining competitive accuracy. Thus, they propose ThunderNet in which a variant of ShuffleNet, dubbed SNet, is proposed to implement an efficient DNN for object detection. 
EfficientDet \cite{tan2020efficientdet} is the counterpart of EfficientNet in object detection, where a new bi-directional feature pyramid network is proposed to combine with EfficientNet to generate an efficient object detection detector. 

\noindent \textbf{Discussion:}  Table \ref{tab:my_label} summarizes all object detection works discussed in this section. Although object detection can be considered as an extension of classification, there lack adequate efforts to address the efficiency issues of object detection for resource-constrained systems. The two-stages methods provide high accuracy at the cost of efficiency, and even on powerful GPUs they cannot guarantee real-time constraints (e.g., 25fps for video). The one-stage approaches can trade off accuracy for efficiency, but the practical deployment on edge devices is still behind the expected performance \cite{zhao2019object}. 
% Temporal requirements are imperative for some time-sensitive or real-time applications, like, self-driving cars, UVA surveillance, etc. 
In addition, we notice that only two works evaluate their designs on edge settings (on ARM CPU, mobile GPU or low power accelerators) and as pointed in \cite{ma2018shufflenet} using the \textit{indirect} metrics lke FLOPs or MACs cannot directly translate to the relevant performance metric, i.e., latency and throughput. \textit{Therefore, for the future model design, evaluating models' direct performance on targeting platforms will be a necessity. In addition, for edge systems, the trade-off between speed and accuracy should be application or context dependent, so a good benchmark and design guide should be developed for practitioners \cite{huang2017speed}}. 
% The insights learned from handcrafted models are really handy for 
% \di{add something!!!}

\section{Network Compression}
\label{section:compress} 
Novel lightweight models provide us a solution to efficiently deploy DNN models on edge devices. However, designing a novel architecture is really challenging due to its large design space and complicated parameter tuning. 
% The light-weight models account for a small portion of hand-crafted DNNs. 
Unfortunately, the majority of DNN models are designed to pursue better accuracy without consideration of resource constraints of edge systems, where complex DNN models have a few hundred layers and several billions of parameters to achieve competitive accuracy. As indicated in \cite{denil2013predicting}, DNN models usually have significant redundancy in terms of weights and parameters. Then, an interesting question is raised: 
% The highly computational overhead  of these models raises a critical question for their execution efficiency, especially on resource-limited hardware, like mobile phone and edge devices: 
\begin{tcolorbox}
\textit{Can we reduce the complexity of DNN models by removing these redundancy without greatly compromising their predictive performance?}
\end{tcolorbox}

\textit{Network compression} tackles this problem by removing the redundancy of over-parameterized networks.
% The vast methods are proposed to reduce the complexity of DNN models not specifically designed for edge systems. 
Network compression techniques generally fall into three categories: \textit{network pruning} which removes the redundant weights and channels of over-parameterized DNNs, \textit{quantization} which uses fewer bits to store DNN weights and intermediate results (e.g., float point 32, FP32 to integer 8, INT8), and \textit{knowledge distill} which learns a small and compact (student) model from a large and over-parameterized (teacher) model. 
It is worthy noting that the three approaches are not mutual exclusive to each other and in many cases they are combined to maximally compress the redundant models. In this section, we discuss three approaches in details. 
% structured transform \cite{sindhwani2015structured}

\subsection{Network Pruning}
\label{compress:pruning}
% \begin{table*}[]
%     \centering
%     \caption{DNN model compression methods}
%     \begin{tabular}{c|c|c|c | c}
%     \hline
%         Methods & Years & Category & Compression Ratio & Accuracy Loss \\ \hline
%         Deep Compression \cite{han2015deep_compression}&  2016 & non-structured pruning (weight) & 35x & 0%\\
%     \end{tabular}
%     \label{tab:my_label}
% \end{table*}
The main motivation behind network pruning is that DNN models are usually over-parameterized in terms of weights and channels \cite{denil2013predicting}, and eliminating these redundancy within the model can hugely reduce the computational complexity and storage requirement. 
% Network pruning intends to reduce the redundancy of DNN models. 
Many network pruning methods are proposed in the past 5 years, and we like to classify them into two major branches based on the pruned structure: \nsp and \spt. 
\subsubsection{Non-Structure Pruning}
\begin{figure}
    \centering
    \includegraphics[width=\columnwidth]{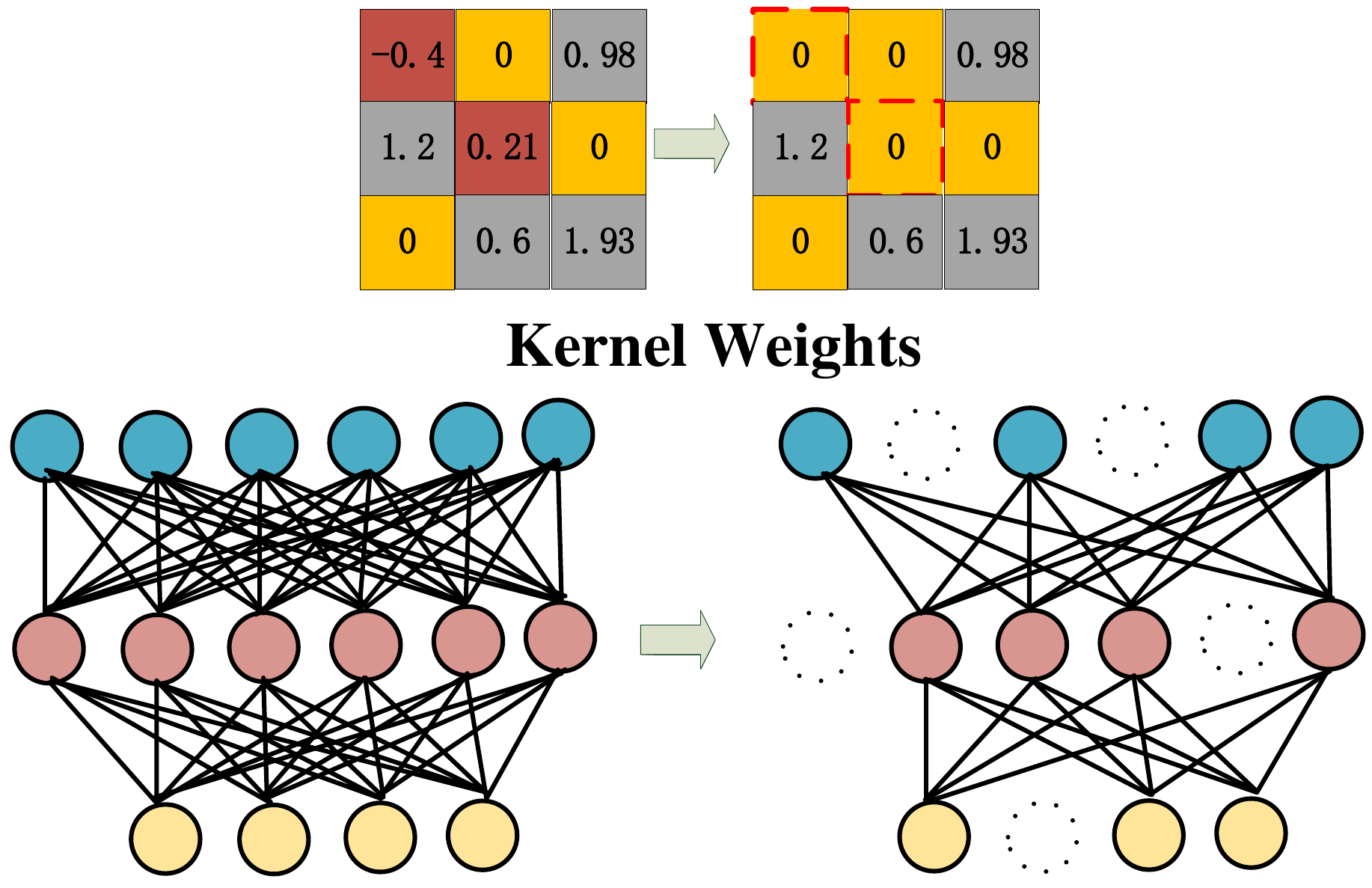}
    \caption{Non-structure Pruning}
    \label{fig:nsp}
\end{figure}

\textit{Non-structured pruning} technique, widely known as \textit{weight pruning}, conducts a fine-grained operation by removing irrelevant weights in over-parameterized DNN models, as shown in Fig. \ref{fig:nsp}. It removes individual weights within a kernel or individual neurons within a fully connected layer. \textit{Non-structure pruning} can significantly reduce the number of parameters and memory footprint. 

Weight pruning on neural network can be traced back to 1990s, \cite{lecun1990optimal,hassibi1993optimal}, where pruning on fully-connected neural networks was investigated.
Deep Compression framework \cite{han2015learning,han2015deep_compression} is one of the pioneers in DNN model compression. Deep Compression uses three steps, \textit{prunning}, \textit{quantization}, and \textit{Huffman Encoding}, to compress over-parameterized DNN models, such as VGG \cite{vgg}, AlexNet \cite{Alexnet}. 
The pruning step removes the weights with magnitude lower than a threshold and corresponding connections. Then, quantization reduces the bit-width of weights to reduce the model size (More details about quantization in the subsequent section). Finally, Huffman encoding further compresses the weight storage. Experimental results show deep compression can significantly reduce the model size with no or negligible accuracy loss. 

Inspired by the success of deep compression, many new methods are proposed to further improve the efficacy of \nsp. 
Molchanov \etal \cite{molchanov2017variational} extend variational drop-rate  to each weight of a DNN model, and weights with high drop-rate are deemed irrelevant and thus can be removed for model compression. 
Different from above-mentioned approaches which conduct \nsp directly on pre-trained networks, NeST \cite{dai2019nest}, that is inspired by the development of human brain, adopts a grow-prune scheme, where NeST first makes a sparse seed DNN model bigger and more complex and then prunes some irrelevant weights from the grown model to generate the final compact model.  
Zhang \etal \cite{Zhang_2018_ECCV} formulate the weight-pruning problem as a non-convex problem which can be solved by \textit{alternating direction method of multipliers} (ADMM) method \cite{boyd2011distributed}.

All methods discussed above target to reduce the model size via \nsp. However, for edge devices, power and energy are also important metrics to consider.
% the model size as the main objective, i.e., they aim to or in literature intend to reduce memory size by removing small magnitude weights and accordingly unused neurons which have no or small impact on predictive accuracy. 
Yang \etal in \cite{yang2017designing} propose an energy-aware pruning method, in which they strive to reduce the energy consumption of DNN models by \nsp.
% weights have high impact on data and target to reducing the model energy consumption as consideration another important design concern for edge/IoTs devices, i.e., energy, and propose energy-efficient weight pruning. 
The core idea behind their approach is to order layers according to their energy consumption and then it prunes weights according to that order.

Although \nsp can significantly reduce memory footprint and multiply-accumulate (MACs) of DNNs, \textit{such reduction does not directly translate to latency improvement}. This is because \textit{non-structured pruning} generates sparse structures which lead to irregular access pattern. The irregular pattern of sparsified DNN models needs special formats, e.g., compressed sparse row and compressed sparse column, to store sparse matrices. The off-the-shelf hardware and software cannot efficiently execute those compressed formats, so specialized hardware and software libraries are required to execute sparsified DNN models \cite{chen2016eyeriss,han2016eie}.

\begin{table}[]
    \centering
    \caption{Comparison of Non-structure pruning methods}
    \label{tab:pruning_nsp}
    \resizebox{0.93\columnwidth}{!}{%
    \begin{tabular}{c|c|c |c}
    \hline
        Methods &  Pruning method & Metrics & year\\ \hline
        Deep Compression \cite{han2015deep_compression} & Threshold/Huffman Coding &  Model size &2015 \\
        Molchanov \etal \cite{molchanov2017variational} & Variational Drop Rate & Model size &2017 \\
        Yang \etal \cite{yang2017designing} & heuristic & Energy &2017 \\
        Zhang \etal \cite{Zhang_2018_ECCV} & ADMM & Model size &2018 \\
        NeST \etal \cite{dai2019nest} & Grow-prune & Model size &2019 \\
        \hline
    \end{tabular}
    }
\end{table}

\subsubsection{Structure Pruning}
\begin{figure}
    \centering
    \includegraphics[width=\columnwidth]{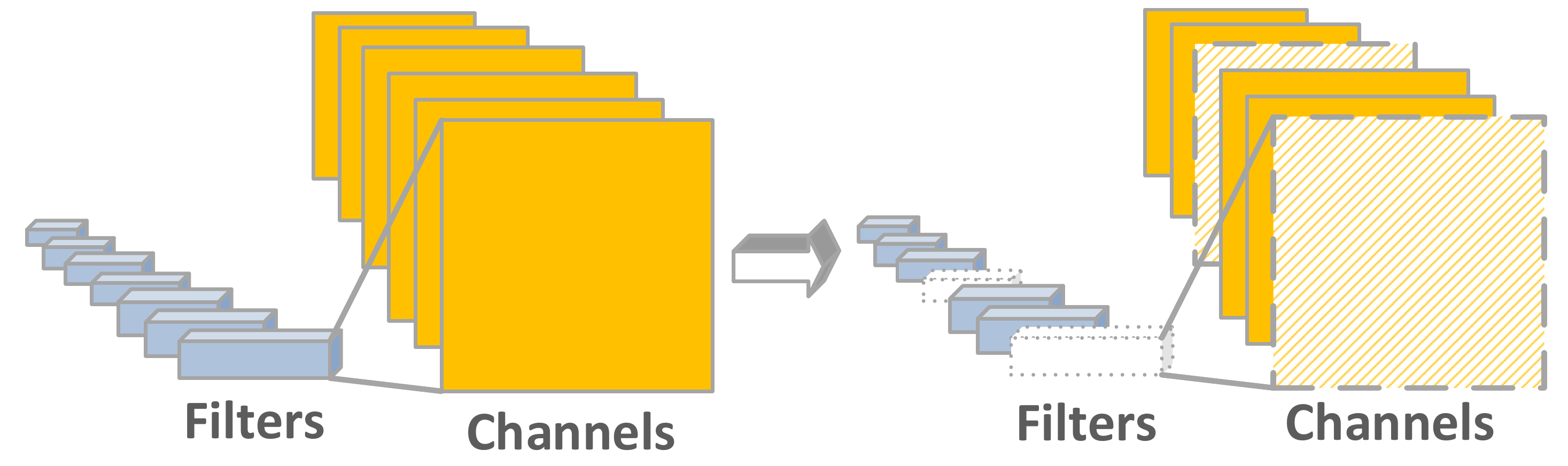}
    \caption{A simple visualization of filter pruning}
    \label{fig:filter_pruning}
\end{figure}

\textit{Structure pruning}, on the other hand, prunes network by maintaining its regular pattern. To keep regularity, \spt completely removes some channels and filters, that have least impact on the model's prediction as shown in Fig. \ref{fig:filter_pruning}. Since \textit{structured pruning} does not lead to irregular pattern, the compressed network pruned by \spt can directly accelerate its inference on off-the-shelf hardware platforms without specialized software library support. Therefore, it has been receiving growing attention in recent years. 

The common process of \spt is (1) \textit{defining a pruning criterion}; (2) \textit{selecting pruned channels according to the criterion and goal}, such as compression ratio and the number of MACs or FLOPs; and (3) \textit{fine-tuning the pruned model}, i.e., retraining the pruned network to retain accuracy.
Works in \spt define different criteria and adopt different methods to select the pruned channels, while minimizing accuracy loss. In terms of pruning methods, we can classify them into two categories: \textit{training-based} vs \textit{inference-based}. 
\begin{itemize}
    \item \textbf{Training-based}: The pruning is conducted during the training procedure, where a sparsity constraint is exposed and a compact network is directly learned from a big and over-parameterized network;
    \item \textbf{Inference-based}: The prune method reduces the redundant channels from a pre-trained model according to defined rules; 
\end{itemize}
In terms of pruning scope, we have \textit{layer pruning} vs \textit{global pruning}. 
\begin{itemize}
    \item \textbf{Layer pruning}: The pruning process is applied to the network layer by layer for finding the pruned network satisfying a defined target; 
    \item \textbf{Global pruning}: The pruning process is applied to the whole network for finding the best pruned network while satisfying a defined target;
\end{itemize}
The pruning process is either \textit{rule-based} or \textit{learning-based}.
% conducted with defined objectives such as  either \textit{manually} or \textit{} and From the algorithmic point of view, the existing \spt can be classified into heuristic algorithm pruning vs automatic pruning. 
\begin{itemize}
    \item \textbf{Rule-based}: The pruning is conducted according to some defined rules, like heuristic algorithms;
    \item \textbf{Learning-based}: The pruning is conducted by a learning algorithm, such as reinforcement learning \cite{sutton2018reinforcement}, evolutionary algorithms \cite{ea} and gradient-based optimization.
\end{itemize}

\begin{table}[]
    \centering
    \caption{The classification of Structure Pruning methods.}
    \resizebox{0.93\columnwidth}{!}{%
    \begin{tabular}{c|c|c|c|c}
     \hline
        Methodologies &  Pruning methods & Scope & Process & Year\\ \hline
        Li \etal \cite{li2016pruning} & Inference-based & Layer & Rule-based & 2016 \\
         Hu \etal \cite{networktriming} & Inference-based & Global & Rule-based &  2016\\
         SSL \cite{wen2016learning}& Training-based & Layer & Rule-based & 2016 \\
        He \etal \cite{he2017channel} & Inference-based & Layer & Rule-based & 2017 \\
         ThiNet \cite{luo2017thinet} & Inference-based & Layer & Rule-based & 2017 \\
          DeepIoT \cite{yao2017deepiot} & Training-based & Layer & Learing-based & 2017 \\
         DCP \etal \cite{zhuang2018discrimination} & Inference-based & Layer & Rule-based & 2018 \\
          SFP \cite{he2018soft} & Training-based & Layer & Rule-based & 2018 \\ 
          Huang \etal \cite{huang2018data} & Training-based & Global & Rule-based & 2018 \\
       AMC \cite{He_2018_ECCV} & Inference-based & Layer & Learning-based & 2018 \\ 
         NetAdapt \cite{yang2018netadapt} & Inference-based & Layer & Learning-based & 2018  \\ 
         Gate Decorator \cite{you2019gate} & Inference-based & Global & Rule-based & 2019 \\
         LFPC \cite{he2020learning} & Inference-based & Layer & Learning-based & 2020 \\
          HRank \cite{hrank} & Inference-based & Layer & Rule-based & 2020 \\\hline
    \end{tabular}
    }
    \label{tab:spt}
\end{table}

Li \etal \cite{li2016pruning} adopt a global pruning method where all filters are sorted in terms of absolute weight sum and then filters with low magnitude are pruned and related channels are all removed.  
% the importance of filters within each layer by calculating the weight sum of each filter and prune the unimportant filters to reduce the model complexity. 
% prune filters inside a network based on value magnitude of each filter, which is analogous to the weight magnitude pruning discussed in the previous section, where the filters with small value magnitude are pruned to reduce the complexity.   
% Pruning layers may damage the structure of the subsequent layer, so reconstructing the damaged layers are needed. 
% If some filters are pruned, they will affect the structure of the subsequent layers. 
He \etal \cite{he2017channel} present a layer pruning method using LASSO regression and reconstruction error to select pruning channels. 
Hu \etal \cite{networktriming} observe that a fraction of activation weights in DNNs are zero and these zero weights imply the corresponding filters are likely to be redundant and can be pruned, and they thus propose using Average Percentage of Zeros (APoZ) of a filter as the criteria to select pruned channels.
Instead of using the information from  the currently pruned layer for selecting pruned channels, ThiNet \cite{luo2017thinet} proposes to exploit the information from the output of the next layer to determine pruned filters.  
% by minimizing the reconstruction error of the output feature map in a single layer, which discards those channels and corresponding filters which have less impact on the output feature map in each layer to cut down the size of the whole model.
% This approach can be applied on both single-layer pruning and the whole model compression.
Discriminate-aware channel pruning (DCP) in \cite{zhuang2018discrimination} relies on the discrimination of each filter to select the pruned channels. The main concept is that the discriminated channels provides more relevant information or features to retain accuracy and then the channels which are inadequately discriminated can be pruned for complexity reduction.  
You \etal \cite{you2019gate} propose \textit{gate decorator} module to replace the batch normalization module in DNNs to select pruned filters globally.  
Most \spt methods adopt a uniform pruning criterion for all layers, but different layers have different functions, thereby likely benefiting from employing different pruning criteria at different layers. Recently, He \etal \cite{he2020learning} propose LFPC to learn an optimal pruning criterion for each layer by using a gradient-based method. HRank \cite{hrank} empirically finds that filters with low-rank is less informative than those with high-rank, and thus uses this observation to prune the unimportant channels.

Wen \etal \cite{wen2016learning} propose SSL to have a training-based pruning method, learning a compact and sparse model from a pre-trained model. 
Liu \etal \cite{networkslimming} propose an approach called network slimming, which takes wide and large networks as input models, but during training,  insignificant channels are automatically identified and pruned afterwards.
Soft filter pruning (SFP) \cite{he2018soft} deploy a training-based pruning method to prune a complex network. 
Huang \etal \cite{huang2018data} propose the concept of sparsity scaling factor for each filter which is learned during to training, and then filters with scaling factor 0 are removed. 
Yao \etal \cite{yao2017deepiot} train an recurrent neural network (RNN) to determine the dropout probability of each filters and prune the network layer by layer according to drop probability.

% Ren \etal \cite{ren2019darb} propose a density-adaptive regular-block pruning for deep neural networks.

Most of the above-mentioned \spt methods are rule-based, i.e., some heuristic algorithms devised according to their own criterion. In contrary, some work employ learning algorithm to automatically prune network model. 
AMC \cite{He_2018_ECCV} proposes to use \textit{reinforcement learning} to automatically prune channels of each layer. 
Anwar \etal \cite{anwar2017structured} prune a DNN model at feature level, kernel level, and intra-kernel level (i.e., weight), where they deploy \textit{evolutionary algorithm} to find the best combination of different pruning granularities. 
However, searching in a discrete space using RL and EA is really costly, so the gradient-based method is recently proposed to find an optimal pruning criterion for each layer \cite{he2020learning}.
% He \etal \cite{he2020learning} recently propose to learn filter pruning criteria (LFPC) to prune the network. 

%slimmable network improvement \cite{yu2019universally}

% Network pruning tends to trim the vast complicated models such that they can be efficiently deployed on resource constrained devices. Nowadays, 

To determine how many channels should be pruned, the above-mentioned works use indirect metrics like FLOPs or compression ratio to prune network. Nevertheless, the reduced FLOPs and compression ratio cannot directly translate to the performance improvement. In addition, a diverse of hardware accelerators have emerged for boosting the execution of DNNs, but these various systems demonstrate different capability to handling network complication. 
Hence, some pruning studies directly target the direct metric upon a specific hardware, e.g., latency. NetAdapt \cite{yang2018netadapt} proposes an automated framework to prune filters in different layers such that the pruned model can be adapted to a target platform. To optimize the latency on a target platform, NetAdapt builds up a look-up-table (LUT) for different operations and layers, so instead of measuring latency on the real platform, it can quickly estimate the latency based on the model architecture and LUT. 
Yu \etal \cite{Yu2017ScalpelCD} introduce SIMD-aware pruning framework which employs different pruning strategies for different underlying hardware, like weight pruning for low-parallelism CPU and filter pruning for high-parallelism.

\noindent{\bf Discussion:}
Table \ref{tab:pruning_nsp} and \ref{tab:spt} summarize methods discussed in this section. Pruning is the well-studied topic in model compression and many methods have been proposed to advance the network pruning \cite{blalock2020state}. These pruning methods show promising results, capable of compressing a DNN model, in some cases more than 40x, while only degrading the prediction accuracy slightly \cite{han2015deep_compression}. 
To effectively adopt model pruning for edge intelligence systems, the existing works suffer from two flaws: 1) 
the theoretical foundation behind pruning lacks, and thus there is a debate whether we should prune a complex model or directly train a compact model for resource constrained hardware platforms. Some recent research \cite{liu2018rethinking}\cite{frankle2018the}\cite{wang2019pruning}\cite{malach2020proving} strives to empirically or theoretically find an answer for this question. 
Since pruning is a time-consuming procedure involving large model training and iterative pruning-retraining procedure, a theoretical foundation or proof may drastically change the way we use model pruning to design edge DNN models; 
2) Almost all of pruning methods are \textit{hardware-agnostic} and depend on \textit{hardware-independent} metrics, like MACs and FLOPs. Since different hardware architectures demonstrate different degrees of parallelism, the state-of-the-art methods may unnecessarily prune models without performance improvement (e.g., latency reduction) while reducing the capacity of DNN models which is proven to have a significant impact on model's accuracy \cite{wideresnet}. Therefore, we need to devise pruning method with hardware-awareness such that the model can be tailored for various hardware. 
% In addition, many design methods like \cite{Cai2020Once} train a supernet (an overparameterized network) and sample a subnet from the supernet based on the hardware constraints and performance requirements, and some methods \cite{fedorov2019sparse} integrate pruning into the pipeline of designing compact DNN models.

% for example, 

% \textit{If a pruned, compact model can achieve the same accuracy as its original, complex model, why can not directly train a small model from scratch?}

% At beginning, it was believed that a compact network directly trained from scratch can not achieve the same accuracy as the one pruned from an over-parameterized network and the weights inherited from the large network model can help the pruned model to retain its accuracy \cite{han2015deep_compression}\cite{li2016pruning}. 
% However, some recent research \cite{liu2018rethinking}\cite{frankle2018the}\cite{wang2019pruning} shows the contrary opinion and experimental evidence that training a dense network from scratch is actually able to achieve comparable accuracy as their pruned counterpart. 

\begin{figure}[]
     \centering
         \includegraphics[width=0.95\linewidth]{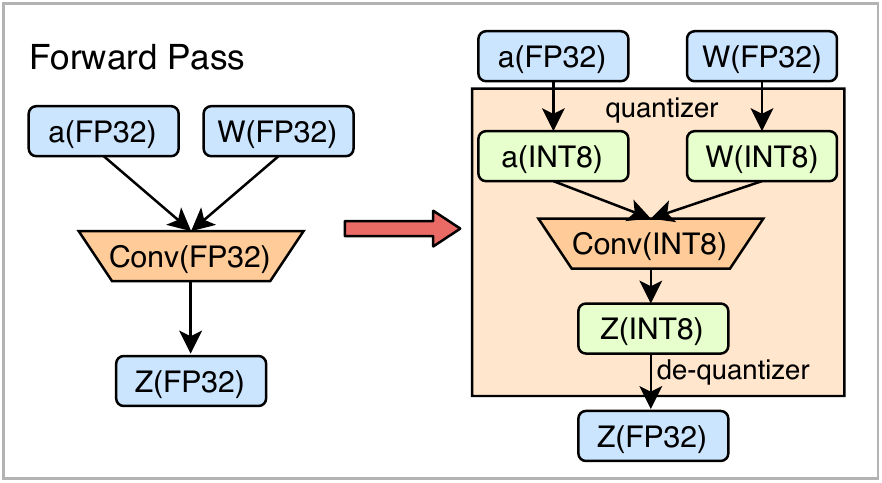}
     \caption{An example of convolution quantization.}
     \label{fig:quantization}
\end{figure}

\begin{table}[t]
    \centering
    \caption{The classification of quantization methods.}
    \resizebox{0.93\columnwidth}{!}{%
    \begin{tabular}{c|c|c|c|c}
     \hline
        Methodologies &  Bits & Weights & Activations & Year \\ \hline
        Gupta \etal \cite{gupta2015deep} & 16-bit & Yes & No &  2015 \\
        Q-CNN \cite{wu2016quantized} & 1-bit & Yes & No & 2016  \\
        BinaryConnect \cite{binaryconnect} & 1-bit & Yes & No & 2016  \\
        Ternary \cite{ternary} & 2-bit & Yes & No & 2016\\
        DoReFa-Net \cite{zhou2016dorefa} & Arbitrary & Yes & Yes & 2016 \\
        XNOR-Net \cite{xnornet} & 1-bit & Yes & Yes & 2016 \\
        INQ \cite{zhou2017incremental} & 5/4-bit$^{\dagger}$ & Yes & Yes & 2017 \\
        ABC-Net \cite{lin2017towards} & 1-bit & Yes & Yes & 2017 \\
        Zhu \etal \cite{trainedternary} & 2-bit & Yes & No & 2017 \\
        Jacod \cite{jacob2018quantization} & 8-bit & Yes & Yes & 2018 \\
        Bi-Real-Net \cite{liu2018bi} & 1-bit & Yes & Yes & 2018 \\
        LQ-Net \cite{zhang2018lq} & Arbitrary & Yes & Yes & 2018 \\
        TQT \cite{jain2019trained} & 8-bit & Yes & Yes & 2019 \\
        HAQ \cite{Wang_2019_CVPR} & (1$\sim$8)-bit$^{\mathsection}$ & Yes & Yes & 2019 \\
        HAWQ \cite{dong2019hawq}  & 2/4-bit$^{\dagger}$ & Yes & Yes & 2019 \\
        Jung \etal \cite{jung2019learning} & (2$\sim$4)-bit & Yes & Yes & 2019 \\
        Zhuang \etal \cite{zhuang2020training} & 2-bit & Yes & Yes & 2020 \\
        XOR-Net \cite{XOR-Net} & 2-bit & Yes & Yes & 2020 \\
    \hline
    \end{tabular}
    }
    $a$/$b$-bit $^{\dagger}$ denotes the method quantizes the network weights for $a$ bits while operating intermediate activation feature maps for $b$ bits. \\ 
    ($a \sim b$)-bit$^{\mathsection}$ represents the method quantizes the network weights or activations from $a$ bits to $b$ bits.
    \label{tab:quantization}
\end{table}

\subsection{Quantization}
\label{compress:quant}
Network pruning reduces the complexity of DNN models by removing redundant weights or channels. 
However, the state-of-the-art models have more than billions of parameters and at the same time during inference a model produces a large portion of intermediate results (activation/feature maps) which usually occupy a large memory space. 
As a result, the huge memory requirement prohibits DNN models from implementing on memory-limited edge devices \cite{lane2015early,tinytransfer}. 
For example, ResNet-50 \cite{he2016deep} has 26 million parameter weights, generates 16 million activations in one inference, requires around 168 MB memory space, and needs at lease 3GB/s memory bandwidth.  
It is not difficult to see that it is unlikely to deploy these state-of-the-art models to edge devices, which have limited storage and computational resources. 

In this case, \textit{quantization} becomes a promising approach to address the aforementioned issue, which encodes full-precision (FP32) weights and activations with low-precision ones (e.g., FP16, INT8, binary) while preserving even the same level of accuracy.
Some early work has shown that using FP16 to train DNN models can reduce the computational cost while retaining accuracy \cite{gupta2015deep}. 
Quantization significantly benefits DNN models on resource-limited devices, and it is capable of fitting the whole model into on-chip memory of edge devices such that the high overhead occurred by off-chip memory access can be mitigated. In addition, since operations with low-bit representation usually consume less energy and execute faster, quantization reduces energy consumption and latency as well on some hardware platforms \cite{dally2020domain}. 
In this section, we discuss some state-of-the-art quantization studies.

Wu \etal \cite{wu2016quantized} propose a unified quantization framework, which improves the quantization performance by minimizing the estimation error of each layer's response. On this basis, an error correction training strategy is included. 
Jain \etal \cite{jain2019trained} propose a trained uniform quantization method for accurate and efficient neural network inference on fixed-point hardware. 
Jacob \etal \cite{jacob2018quantization} quantize models into 8-bit integer and propose a quantize-aware training method to eliminate the accuracy loss caused by quantization. Instead of quantizing all weights in a DNN, IQN \cite{zhou2017incremental} adopts a group-wise quantization method to gradually quantize weights of a DNN model. The advantage of IQN is that it is able to derive the quantized model without accuracy loss.  

The above works uniformly convert all weights and activation into the same low-bit representation, but many emergent hardware and accelerators support mixed precision operations, e.g., Nvidia Turing GPU architecture supports 1-bit, 4-bit, 8-bit and 16-bit arithmetic operations \cite{nvidia}. This provides a more effective and flexible way to quantize weights and activations of a DNN model.  
% focus on the algorithm level, which do not consider the real-world situations, which ignore the interactions with hardware devices. Benefitting from the fact that deep learning accelerators begin to support mixed precision operations, 
Wang \etal \cite{Wang_2019_CVPR} propose a hardware-aware automated quantization approach, namely HAQ. HAQ exploits reinforcement learning to select different quantization width for each layer upon a target hardware. 
Additionally, the hardware architecture is involved into the learning loop, so that it can directly reduce the inference latency, energy and storage on the target hardware. HAWQ \cite{dong2019hawq} also considers to quantize a full-precision model into a mix-precision model where the quantized precision is determined for each layer based on Hessian matrix.  
% To this end, HAQ model the quantization task as a reinforcement learning problem, which apply the actor-critic model with DDPG agent to give the action: bits for each layer. Additionally, hardware counters together with the accuracy are collected to formulate the optimization objective. 

% \cite{zhang2018lq}

% \cite{jung2019learning}

Some works use binary or ternary quantization to maximally compress DNN models. Then, binarized or ternarized network can use cheap bit operation to boost the efficiency on dedicated hardware \cite{umuroglu2017finn}. 
In \cite{binaryconnect}, Courbariaux \etal propose the binaryconnect, which targets to transform the full-precision weights into the binary format. A very straightforward binarization method would be based on the sign function:
\begin{equation}
    w_b = \begin{cases}
     +1 \,\, \ifff w \ge 0\\
    -1 \,\, \otherwise 
    \end{cases}
\end{equation}
where $w_b$ is the binary weight and $w$ is the full-precision weight. Due to the fact that this is a deterministic operation, averaging the discretization over the many input weights of a hidden unit could compensate for the loss of information. An alternative that allows a finer and more correct averaging process to take place is to binarize stochatically, which helps to improve the model generalization capability:
\begin{equation}
w_b = \begin{cases}
    +1 \,\, \with \, \possibility p = \delta(w) \\
    -1 \,\, \with \, \possibility 1 - p
\end{cases}
\end{equation}
where $\delta$ is the \textit{hard sigmoid} formula:
\begin{equation}
    \centering
    \delta(x) = \clip (\frac{x+1}{2}, 0, 1) = \max(0, \min(1, \frac{x+1}{2}))
\end{equation}

Intuitively, applying a binarized method is an easy way to quantize the full precision weights. However, this will be harder for the training process to converge due to the highly discrete parameter space, which drastically degrades the model performance.

Later, Rastegari \etal \cite{xnornet} present an efficient approximation strategy, which constrains the full precision weights to $+1$ and $-1$ instead of directly rounding them. With a scaling factor $\alpha$, an convolutional operation can be approximated as follows:
\begin{equation}
    \centering
    \mathcal{I} * \mathcal{W} \approx \mathcal{I} * (\alpha \mathcal{B}) = \alpha \mathcal{I} * \mathcal{B}
    \label{eq:quantization}
\end{equation}
where $\mathcal{I}$ is the input image and $\mathcal{B} \in \{-1, +1\}$ is the binarized weight. Other binarized methods are proposed to improve the degraded accuracy caused by binarization like \cite{zhou2016dorefa, lin2017towards, liu2018bi}. XOR-Net \cite{XOR-Net} takes the implementation of BNNs into account and simplifies the number of instructions used in binary dot product to accelerate BNN inference on edge devices.

Inspired by \cite{xnornet}, Li \etal \cite{ternary} propose ternary weight networks (TWNs) with weights constrained to $\{-1, 0, +1\}$, which minimizes the Euclidian distance between full precision weights $\mathcal{W}$ and the ternary weights $\mathcal{W}^t$ along with a scaling factor $\alpha$. Here the quantized weights are obtained with a threshold-based ternary function:
\begin{equation}
    \mathcal{W}_i^t = \begin{cases}
        +1, \,\,\, \ifff \mathcal{W}_i > \Delta \\ 
        +0, \,\,\, \ifff |\mathcal{W}_i| \le \Delta \\ 
        +0, \,\,\, \ifff \mathcal{W}_i < -\Delta 
    \end{cases}
\end{equation}
where $\Delta$ is a positive threshold parameter. Thus, the optimization objective can be formulated as follows:
\begin{equation}
    \alpha^{*}, \mathcal{W}^{t*} =  \argmin J(\alpha, \mathcal{W}^t) = ||\mathcal{W} - \alpha \mathcal{W}^t||_2^2
\end{equation}
By addressing the above convex optimization problem, we can obtain the approximately optimal $\alpha^*$ and $\Delta^*$:
\begin{equation}
    \begin{cases}
        \alpha_{\Delta}^* = \frac{1}{I_{\Delta}} \sum_{i \in I_{\Delta}}|W_i|, \\
        \Delta^* = \argmax \frac{1}{|I_{\Delta}|}(\sum_{i \in I_{\Delta}}) |W_i|^2 
    \end{cases}
\end{equation}
where $I_{\Delta}$ denotes the number of elements in $\{i | |\mathcal{W}_i| > \Delta\}$. On top of TWNs, Zhu \etal \cite{trainedternary} further introduce two independent quantization scaling factors for positive and negative weights in each layer,  respectively, to improve accuracy of ternary quantization.

Quantization are usually achieved via post-training quantization which is regarded as the most prevalent method currently. The mainstream DNN frameworks like Pytorch \cite{pytorch} and Tensorflow \cite{tensorflow} all support post-training quantization. Fig. \ref{fig:quantization} illustrates the procedure of post-training quantization. First, we should find a proper encoding algorithm, which quantize both full-precision (e.g., FP32) activation results $a$ and networks weights $w$ into the low-precision (e.g., INT8). Then, the operation (e.g., Convolution) is performed as usual, where the derived outputs will be further relaxed to the full-precision format which can be conducted together with the scaling factor (e.g., $\alpha$ in Eq. \ref{eq:quantization}) with respect the quantizer. Post-training quantization is flexible and can be broadly applied to existing DNN models. However, since the weights and activations are quantized into discrete values, we cannot use stochastic gradient descent to update the weights. Consequently, the quantized models may suffer from significant accuracy loss. Therefore, some studies \cite{zhang2018lq, jung2019learning, zhuang2020training} aim to effectively train a low-precision, compact model. Zhang \etal \cite{zhuang2020training} adopt an auxiliary full-precision model to facilitate the training of its quantized counterparts. 

Recently, the robustness of DNN models, i.e., robust to adversarial examples, is a burgeoning topic in DNN research \cite{madry2017towards}. For edge systems, robustness is a critical metric especially for some safety-critical systems, like self-driving cars, UAVs, robots, etc. The failure or mispredication may lead to catastrophic results. Recently, Lin \etal \cite{lin2019defensive} study the effect of quantization on the DNN robustness and propose the Defensive Quantization (DQ) that addresses the robust issue of quantized models, where DQ can maintain adversarial robustness and model performance at the same time.  
Moreover, Gong \etal \cite{gong2019mixed} consider to quantize network with the objective of reducing energy consumption. 

\noindent\textbf{Discussion:} 
Table \ref{tab:quantization} summarizes the quantization approaches discussed in this section. Quantization has become a standard means to compress memory-hungry DNN models for resource-constrained edge systems. Many vendors develop their own tools and hardware to effectively quantize models and support efficient execution of quantized models, like Nvidia TensorRT\footnote{https://developer.nvidia.com/tensorrt}, Tensorflow Lite\footnote{https://www.tensorflow.org/lite}, OpenVino\footnote{https://software.intel.com/content/www/us/en/develop/tools/openvino-toolkit.html}, etc, and some real-world applications based on quantized models are emerging such as \cite{chen2020gpu}. For edge intelligence systems, we have seen a new trend that quantization will work with other model compression techniques as well as the new NAS methods to derive the most efficient and compact models, for example, AQP \cite{AQP} which combines NAS, quantization, and pruning to design efficient DNNs. 
Such method will become a new standard to design edge DNN models.

\begin{table}[t]
    \centering
    \caption{The classification of Knowledge Distillation Methods}
    \resizebox{0.95\columnwidth}{!}{%
    \begin{tabular}{c|c|c|c}
     \hline
        Methodologies &  Distillation Transfer & Number of Teachers & Year \\ \hline
        Remero \etal \cite{romero2014fitnets} & From intermediates & Single & 2014 \\ 
        Hinton \etal \cite{hinton2015distilling} & From logits$^{\dagger}$ & Single & 2015 \\ 
        Zagoruyko \etal \cite{zagoruyko2016paying} & From intermediates & Single & 2016 \\
        Tarvainen \etal \cite{tarvainen2017mean} & From logits & Multiple & 2017 \\
        Polino \etal \cite{polino2018model} & From logits & Single & 2018 \\
        Ravi \etal \cite{ravi2019efficient} & From logits & Single & 2019 \\
        Li \etal \cite{li2020few} & From logits & Single & 2020 \\
        Chung \etal \cite{chung2020feature} & From intermediates & Single & 2020 \\
        Liu \etal \cite{liu2020adaptive} & From intermediates & Multiple & 2020 \\
    \hline
    \end{tabular}
    }
    
    $^{\dagger}$ Following \cite{hinton2015distilling}, we use logits to denote the knowledge from the output of the neural network, \textit{i.e.}, from the last layer.
    \label{tab:distillation}
\end{table}

\subsection{Knowledge Distillation}
\label{compress:kd}
%\textit{Knowledge distilling} is another way to derive a compact model, where a compact student model can `\textit{learn}' the knowledge from a large teacher model. Hinton \etal \cite{hinton2015distilling} propose a method to improve the acoustic model of a heavily used commercial system by distilling the knowledge in an ensemble of models into a single model.
\textit{Knowledge distillation} is another technique to conduct model compression, where a more compact student model can learn the knowledge from a complicated and powerful teacher model. Bucila \etal \cite{bucilu2006model} first propose the concept of knowledge distillation, and Hinton \etal \cite{hinton2015distilling} generalize knowledge distillation and apply it to DNNs. 
% Fig. \ref{fig:knowledge-distillation} shows how knowledge distillation works. 

\begin{figure}[]
     \centering
         \includegraphics[width=0.95\linewidth]{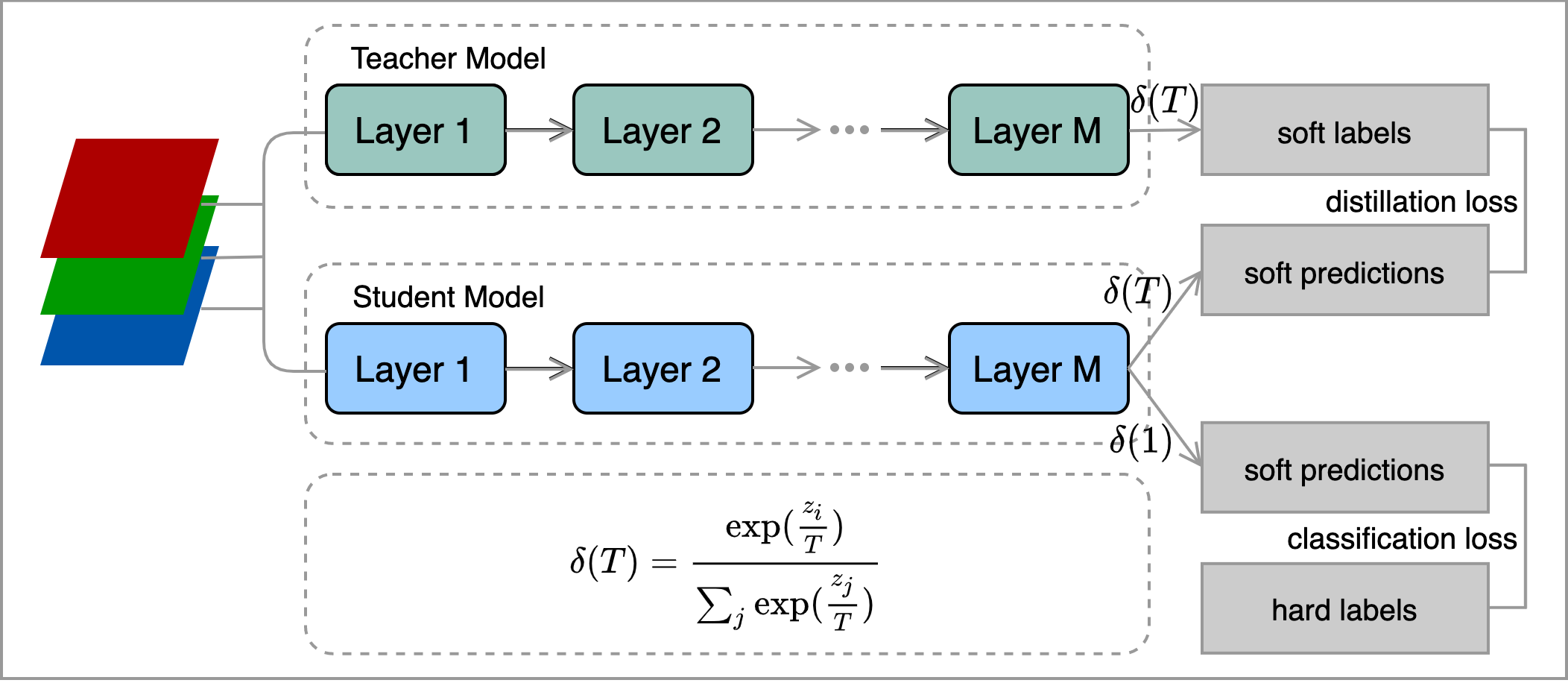}
     \caption{Overview of Knowledge Distill.  }
     \label{fig:knowledge-distillation}
\end{figure}

The core idea of knowledge distillation is to train a compact model (student) with the assistant of a complicated, pre-trained model (teacher).
% In knowledge distillation, there exist a teacher model and a student model. During training, the student model receives not only the information from the training dataset but also the knowledge transferred from the teacher model. 
During training, the student model exploit the conventional method to train the model and obtain a loss according to the one-hot class distribution, e.g., $[0,0,1,0]$, namely \textit{hard targets} and at the same time the knowledge from the teacher model is distilled and transferred to the student model by calculating a new loss in which the target is the probability distribution of predicted class $\mathcal{P}$ from the teacher model, e.g., $[0.1,0.21,0.6,0.09]$, namely \textit{soft target}. Nevertheless, the probability of the correct class dominates the probability distribution generated by the teacher network (e.g., $[0.97, 0.1, 0.0, 0.2]$), which significantly limits the knowledge transferring capability. To alleviate this issue, Hinton \etal \cite{hinton2015distilling} propose \textit{softmax temperature} in which temperature $T$ is to soften the generated probability distribution. Intuitively, a larger $T$ leads to a `softer' probability distribution (e.g., $[0.4, 0.2, 0.2, 0.2]$). Hence, we are able to formulate the softmax with temperature as follows:
\begin{equation}
    \centering
    \mathcal{P} = \{p_i | \frac{\exp(\frac{z_i}{T})}{\sum_j \exp(\frac{z_j}{T})} \}
\end{equation}
when $T$ is set to 1, it is the original softmax function. Please note that we refer the softmax function with temperature $T$  as $\delta_{T}$ for simplicity. Therefore, we can formulate the overall loss function as:
\begin{equation}
    \mathcal{L}(x;W) = \alpha * \mathcal{F}(y, \delta_1(z_s)) + \beta * \mathcal{F}(\delta_T(z_t), \delta_T(z_s))
\end{equation}
where $\mathcal{F}$ denotes the cross-entropy function. $\alpha$ and $\beta$ are two balancing factors. $z_s$ and $z_t$ represent the output logits from student model and teacher model, respectively. $y$ is the ground truth. An illustration about how knowledge distillation works is shown in Fig. \ref{fig:knowledge-distillation}. 
% In knowledge distillation tasks, there exist both teacher model and student model. During the training phase, the student model receives not only the information from the training dataset but also the knowledge transferred from the teacher model. 

After \cite{hinton2015distilling}, many efforts are made towards improving the performance of knowledge distillation. The work in \etal \cite{tarvainen2017mean, liu2020adaptive} extend the number of teacher models from one to multiple. 
However, there exist performance difference among those teacher models. To tackle this issue, they propose to assign different weights for each teacher models, and then weighted-average probability distributions from different teachers are applied to supervise the student model. 
Combined with quantization, Polino \etal \cite{polino2018model} introduce the \textit{quantized distillation}, which leverages distillation during the training process by incorporating knowledge distillation loss. 
Ravi \cite{ravi2019efficient} introduces a neural projection approach to design and train efficient on-device neural networks. Preceding to the prediction, input instances are transformed into binary representations, which significantly reduces the memory consumption. 
Afterwards, the prediction weights are learned by knowledge distillation to achieve higher generalization capability. 
In \cite{li2020few}, Li \etal propose to use knowledge distillation to efficiently compress models, where the uncompressed model and compressed model are considered as a teacher-student pair. 
This new method can avoid the time-consuming fine-tuning after pruning and achieve data efficiency. 

The above works only use the knowledge from the outputs of of the last layer in the teacher model. \textit{Can the intermediate knowledge help to obtain a better model?} Remero \etal \cite{romero2014fitnets} adopt knowledge distillation to train a compact model, namely FitNets. The main idea in FitNets is to train a deeper and thinner student with the knowledge transferred from the shallower and wider teacher model. Different from the previous works, the knowledge in FitNets is not only from the final outputs but also from intermediate feature representations of the teacher model. By doing so, the student model in FitNets mimics or imitates the teacher model from different granularity levels. Similarly, Zagoruyko \etal \cite{zagoruyko2016paying} introduce the attention transfer strategy to mimic the attention maps of a powerful teacher network, which proves to improve the performance of the student network.

\noindent\textbf{Discussion:} 
Table \ref{tab:distillation} summarizes the works discussed in this section. As identified in literature \cite{yim2017gift}, knowledge distillation provides several benefits for small network models. 
\textit{Accuracy:} distilling knowledge from large networks can improve the accuracy of small models which may be directly applicable to edge systems. 
\textit{Transferability \cite{pan2009survey}:} Knowledge distillation demonstrates better transferability for the small models, i.e., for a given dataset, learning the small model from knowledge distillation outperforms learning it from scratch. 
This feature is really instrumental to design lightweight models for edge systems, because for some edge systems installed in special 
contexts, there lacks a huge amount of high-quality data to train an accurate model. Thus, 
knowledge distillation facilitates training of a competitive, compact model with small dataset. Moreover, knowledge distillation can also be used to design robust networks \cite{papernot2016distillation}. Knowledge distillation shows several benefits for edge systems and we envision knowledge distillation will be gradually integrated with other techniques like NAS to derive accurate, compact networks.  

%Slimmable Network \cite{yu2019universally}\cite{yu2018slimmable} is similar to \textit{width multiplers} presented in MobileNet V1. Slimmable Network takes as input a pre-trained model, prunes the model with different scaling factors ($<=1$), and outputs a series of models with different trade-off between accuracy and latency.  

%Wang \etal \cite{wang2018skipnet} skipnet

%\cite{hua2019channel}

\section{Hardware-Aware Neural Architecture Search}
\label{section:nas}

\begin{table*}[]
    \centering
    \caption{Comparison of different hardware-aware NAS approaches. Here, cost refers to the time spent to search the network, while we use \textit{GPU Hour} as unit.}
    \begin{tabular}{c|c | c | c | c | c | c }
    \hline
         &  Hardware Evaluation & Search method & Cost (GPU days) & Metrics & Target & Edge Systems\\ \hline
      MnasNet \cite{tan2019mnasnet}  & Measure & RL & 40000 & Latency & Image Classification & Yes\\
      ProxylessNAS \cite{cai2018proxylessnas} & Predictor & gradient & 200 & Latency & Image Classification & Yes\\
      EdgeNAS \cite{edgenas} & Predictor & gradient & - & Latency & Image Classification & Yes \\
      SPNAS \cite{stamoulis2019single} & LUT & gradient & 0.2 $^{\dagger}$  & Latency & Image Classification & Yes \\
      FBNet \cite{wu2019fbnet} & LUT & gradient & - & Latency &Image Classification & Yes \\
      DenseNAS \cite{fang2019densely} & LUT & gradient & 3 &Latency  &Image Classification & Yes \\ \hline
      NAS-FPN \cite{nas-fpn} & - & RL & - & - &Object Detection & Yes*\\
      MnasFPN \cite{chen2019mnasfpn} & Measure & RL & - & Latency & Object Detection & Yes \\
      NAS-FCOS \cite{nas-fcos} & - & RL & - & - & Object Detection & No \\
      Auto-FPN \cite{xu2019auto} & - & gradient & - & - & Object Detection & No** \\
      AdversarialNAS \cite{advnas} & - & gradient & 24 & - &GAN & No \\
      \hline
      SpArSe \cite{fedorov2019sparse} & - & gradient & 24 & Memory &GAN & Yes \\
      OFA \cite{Cai2020Once} & Measure*** & - & - & Mem/Lat & Image Classification & Yes \\
      MCUNet \cite{mcunet} & Measure*** & - & - & Mem/Lat & Image Classifica tion & Yes \\
      \hline
    \end{tabular}
    
    $^{\dagger}$ This low search cost is due to that SPNAS only searches for 8 epochs on a subset of ImageNet.\\
    *Although NAS-FPN is a hardware-agnostic approach, NAS-FPN devises a lite version for resource constrained systems.\\
    ** Auto-FPN incorporate hardware-agnostic resource constraint like FLOPs, which cannot translate the runtime complexity upon target edge systems. \\
    *** These two methods train an over-parameterized and large network which is used to sample different small networks, and the over-parameterized network is designed without consideration of hardware. However, sampled networks can be directly measured on devices. 
    \label{tab:nas}
\end{table*}
Model compression provides a `\textit{large-to-small}' method to generate an efficient and compact model from a complex model for edge devices.
% by reducing redundant weights or channels of over-parameterized DNN models. 
% Model compression takes pretrained model as the seed input and then iterates 
Although it has shown its success in reducing latency and model size, the accuracy of the pruned model is inherently upper bounded by the pre-trained model, i.e., the compressed model cannot have better accuracy than the pre-trained model. Moreover, network compression involves a costly and time-consuming compression-retraining procedure to retain the accuracy of the pruned model. At the same time, as we are witnessing the shift of DNN designs from manual design to automatic search, i.e., neural architecture search, which has demonstrated its capability to design more accurate DNNs without tedious parameter tuning, this immediately raises an appealing question: 

\begin{tcolorbox}
\textit{Can NAS directly design hardware-efficient, accurate neural architectures?} 
\end{tcolorbox}

% However, NAS which use automatic techniques such as reinforcement learning and evolutionary algorithms can find better neural architectures than 

An increasing attention is paid to hardware-aware NAS and some works exploit NAS to design hardware-efficient DNNs.
Tan \etal \cite{tan2019mnasnet} propose a hardware-aware NAS framework, dubbed MnasNet, in which both latency and accuracy are formulated into the reward function of reinforcement learning algorithm and the latency is directly measured from the target mobile device. The reward is shown  in Eq. (\ref{eq:mnasnet}).
\begin{equation}
\label{eq:mnasnet}
    \text{ACC}(m)\times \Bigg[\frac{\text{LAT(m)}}{T}\Bigg]^w
\end{equation}
where $m$ is the obtained network, T is the expected latency and $w$ is a variable to adjust the weight of latency in this reward. ACC and LAT are the real accuracy and latency of network $m$, respectively. 
MnasFPN \cite{chen2019mnasfpn} is an extension of MnasNet which, instead of searching classification model, targets directly searching a network for object detection. 
Dai \etal \cite{dai2019chamnet} present ChamNet which deploys evolution algorithm (EA) and three predictors, i.e., energy predictor, accuracy predictor and latency predictors, to effectively and efficiently search for a DNN model on a target platform. 

The RL-based or EA-based NAS frameworks explore the optimal architecture in a large, discrete search space, so they demand huge amount of computational resource and take thousands of GPU days \cite{liu2018darts} to search for a neural architecture.
As estimated in \cite{cai2018proxylessnas}, MnasNet needs 40,000 GPU hours to search for the optimal network. It hinders users with limited resource to search for a DNN upon their target hardware devices. Thus, some studies aim to reduce the search cost for hardware-efficient NAS.
ProxylessNAS \cite{cai2018proxylessnas} uses a gradient-based method to design hardware-aware neural architectures. Instead of measuring real latency on the target platform, it presents a latency-predictor to facilitate exploration of the optimal configuration for neural architectures. It greatly reduces the search cost to 200 GPU hrs while finding efficient models with competitive accuracy. 
Li \etal \cite{Li_2019_CVPR} consider both latency and accuracy as their NAS objectives and use the concept of \textit{partial order pruning} to reduce the search space such that it can significantly reduce the searching time and achieve a good trade-off between accuracy and latency. 
RCNAS \cite{xiong2019resource} formulates the resource-constrained neural architecture search problem as a submodularity function problem which is known to be NP-Hard but has good heuristic algorithms to approximately solve this problem. 
FBNet \etal \cite{wu2019fbnet} introduces DNAS, which is based on the differential neural architecture search (DARTS) method \cite{liu2018darts}, but instead of just searching for an optimal cell in DARTS, DNAS searches for an optimal setting for each layer within the network. 
DNAS also takes latency as their goal, where a look-up-table is set up for latency prediction. EdgeNAS \cite{edgenas} proposes a novel NAS method to search efficient and competitive DNN model for resource constrained edge devices, where a latency predictor is trained from data collected from various architectures on target hardware devices and the latency predictor is integrated into the DARTS-similar NAS framework for efficiently designing DNN models.
Besides, SPNAS \cite{stamoulis2019single} introduces a single-path paradigm, which formulates kernels with different sizes into one \textit{big} kernel. 
% Thus, we only need to mask out the extra part once we intent to use a smaller kernel.
Similar to DNAS, SPNAS constructs a LUT in terms of the runtime latency of different kernels upon target hardware, which will be incorporated into the differentiable loss function (similar to DNAS and EdgeNAS).
All of the above-mentioned approaches mainly search for the best operators for cells or layers where their width and depth are fine-tuned manually. However, as shown in \cite{tan2019efficient}, the width and depth of a DNN have a critical impact on its accuracy and latency. 
Fang \etal \cite{fang2019densely} propose DenseNAS which not only searches for the optimal architectures but also their width and depth configuration.
The successful application of NAS on image classification inspires researchers to explore the potential application of NAS on other CV tasks, like NAS-FPN \cite{nas-fpn} and NAS-FCOS \cite{nas-fcos} for object detection and AdversarialNAS \cite{advnas} for GAN. However, these methods only consider the accuracy and ignore the performance like latency and power consumption, which are critical for edge intelligence systems. Besides, Auto-FPN \cite{xu2019auto} aims to search for a compact FPN with low FLOPs count, but FLOPs cannot necessarily reflect the runtime performance upon target hardware (see Fig. \ref{fig:xavier_flops}). Similar to MnasNet \cite{tan2019mnasnet}, MnasFPN \cite{chen2019mnasfpn} directly measures the runtime latency on target hardware, thereby significantly increasing the search cost.

% a low latency It can optimize over a layer-wise search space and represent the search space by a stochastic super net. (including latency as the constraints) 

Besides latency, other metrics are also considered in hardware-aware NAS frameworks.  
SpArSe \cite{fedorov2019sparse} targets networks which can be fit into micro-controllers which have small memory footprint and less computation capability, where NAS and pruning techniques are combined to design small-memory networks. 
% In addition, it proposes hardware-aware NAS to optimize the architecture of NN for different platforms like CPU, GPU and FPGA.
Cai \etal \cite{Cai2020Once} propose an \textit{once-for-all} (OFA) framework to train a large super network and then sample different size of networks from the super network to fit the different hardware platforms. The advantage of OFA is that it just needs to train once to generate many different DNNs which can be directly applied on diverse hardware platforms, thus greatly reducing the training cost and $\text{CO}_2$ emission. 
Lin \etal  \cite{mcunet} propose MCUNet aiming to design DNN models for microcontrollers. To fit computation-intensive DNN models on microcontollers, MCUNet consists of two key parts, TinyNAS, a NAS framework to search for models satisfying different constraints, such as memory, latency and TinyEngine, an efficient inference library. 
% When deploying network on different hardware platforms, a network can be selected based on its hardware constraint and performance requirement. 

\noindent\textbf{Discussion}: 
 Table \ref{tab:nas} summarizes the works discussed in this section. NAS defines a new paradigm to design DNN models, frees DNN designers from tedious hyperparameter tuning and automates the costly DNN design procedure. Especially, in future edge era, more and more customized DNN models for specific tasks will be developed and implemented on different hardware platforms, e.g., edge systems with intermittent power supply \cite{gobieski2019intelligence}. 
 Nevertheless, for edge intelligence systems, the existing NAS methods suffer from two flaws. First, the majority of existing methods use  \textit{platform-independent} metrics, i.e., FLOPs and MACs as the constraint to design the model. 
 However, the same model demonstrates significant difference on different hardware \cite{aibenchmark}. 
 Therefore, such designs may generate inefficient models for targeting hardware. 
 Second, the existing NAS methods mainly target the models with high accuracy while ignoring other important system metrics, e.g., energy and latency. This leads to the searched model with low efficiency and high energy consumption.  
%  In this case, NAS may facilitate the reduction of enormous engineering efforts and thus shortens the time-to-market. Although we have seen some efforts in this emergent area, hardware-aware NAS is still in its early age. 
 \textit{For edge intelligence systems with diverse tasks and hardware architectures, we need hardware-aware NAS methods to efficiently tailor competitive DNN models for a specific hardware platform. Moreover, to achieve an optimal design for future edge intelligence systems, NAS may need to take advantage of other model compression techniques like network pruning,  quantization, and knowledge distillation, to have a holistic and efficient design paradigm, e.g., \cite{AQP}. 
 }

%  However, NAS which based on reinforcement learning and evolutionary algorithms are extremely costly in terms of resource. Some recent work, DARTS \cite{liu2018darts} and PC-DARTS \cite{xu2019pc}, uses differential method to significantly reduce searching cost. FBNet shows the potential of these methods on hardware-aware NAS, but hardware-aware NAS is still in its early age. To speed up the deployment of DNN on edge systems and customize network for different hardware, we may need to combine NAS with other techniques, like network pruning, and quantization.  
\section{Adaptive Models}
\label{sec:adaptive}

\begin{figure}
    \centering
    \includegraphics[width=\columnwidth]{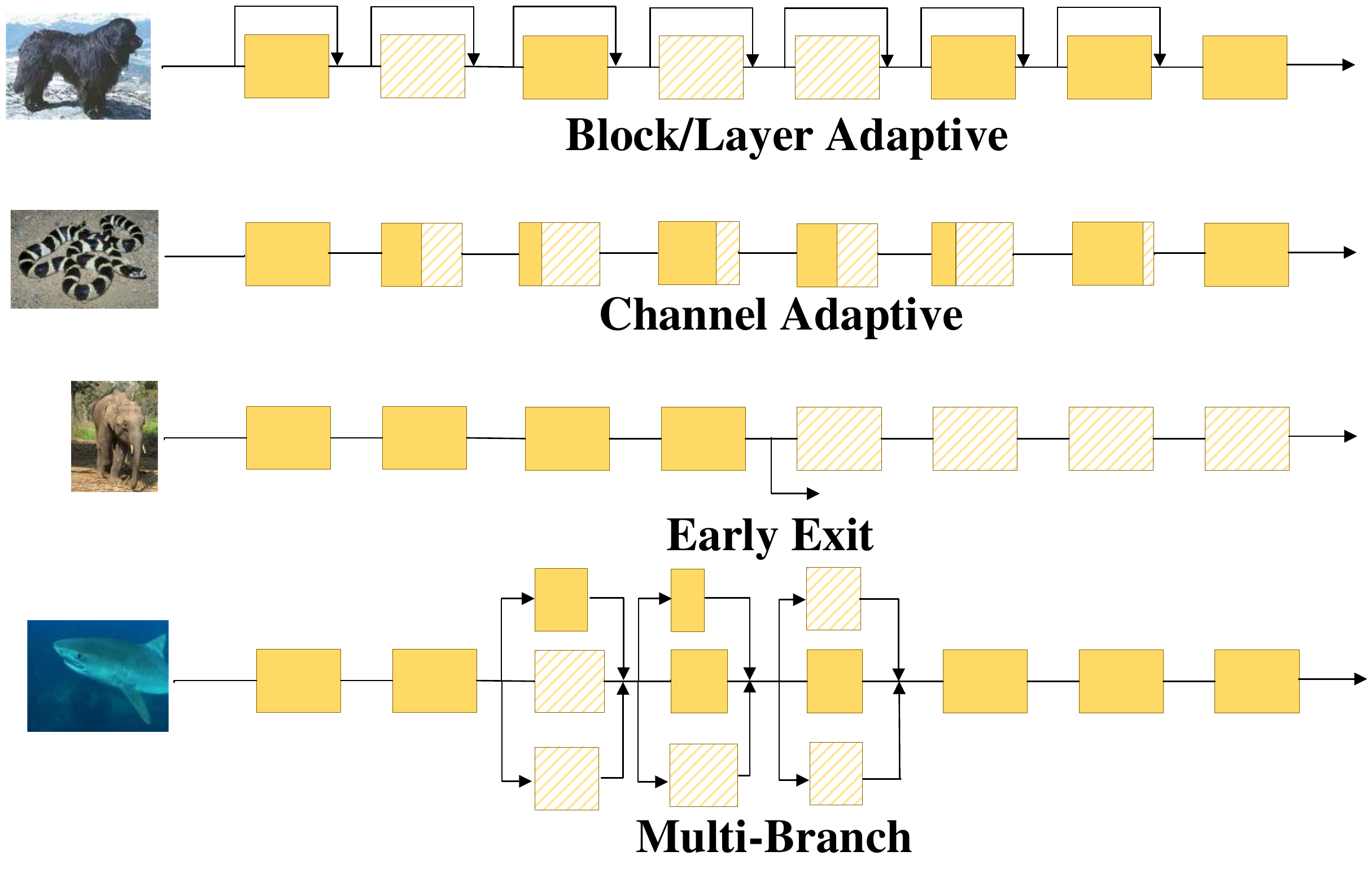}
    \caption{Visualization of four different methods for adaptive models}
    \label{fig:adaptive}
\end{figure}

In previous sections, we discuss the different ways to design lightweight DNN models for edge systems, all of which generate a static DNN model, i.e., the model does not change its execution behavior during run-time. However, during the run-time, 
% DNN models edge systems are installed in diverse contexts, such as communication station, UAVs, Robotics shown as Fig \ref{fig:overview}. 
DNN applications share the computational units and communication bandwidth with other applications and as a result the availability of computing and communication resources may significantly affect DNNs' inference time. 
In addition, some edge systems are energy-constrained, battery powered or supplied by sustainable energy, like solar energy (intermittent computing \cite{intermittent}), and in these cases energy variance will change the system's status, such as scaling down frequency and power which affects the inference time of DNN applications as well.

Such variance may influence the quality of service (QoS) of applications without rigorously temporal requirement, like voice recognition, face recognition, machine translation, etc \cite{facebook} and on the other hand may lead to catastrophic consequence for those with rigorously temporal requirement, (i.e., real-time requirement), like autonomous driving, UAV, etc. Execution variance of DNN applications require DNN models to be adaptive to the different input data and system status (like low power mode) for guaranteeing certain QoS or real-time performance. 
Adaptive DNN models will be useful and practical for edge systems under dynamic environments and prior review articles rarely discuss this topic in a systematical way.  
% \textit{Worth noting that adaptive models indicate the model with a fixed topology but may change the execution path of internal layers to have different computational overheads.}  

\begin{figure}
    \centering
    \includegraphics[width=\columnwidth]{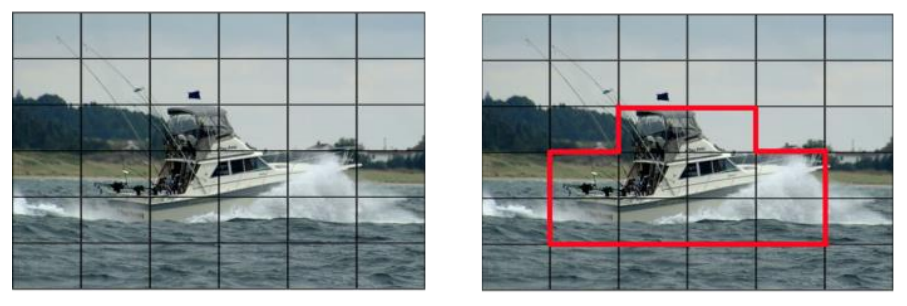}
    \caption{Attention-based methods only conduct convolution operations on `important' areas to reduce the computational cost, where the red line highlights the area of interest.}
    \label{fig:attention}
\end{figure}

We, in this section, review some important techniques of adaptive DNN models which are applicable to edge systems. 
Some early works identify that for different input images, not all channels or layers of a DNN model are needed to make accurate prediction \cite{bengio2013estimating}. 
This key observation serves the technical foundation for adaptive models, i.e., a network may selectively activate its channels and layers per input basis. The partially activated network can achieve higher efficiency than the originally full network without accuracy loss. 
Some initial works on this topic are called \textit{conditional computation} \cite{bengio2013estimating,davis2013low}, where the main objective of \textit{conditional computation} is to enhance the capacity of a network while not significantly increasing the computational cost. However, for edge systems, the efficiency is our top priority, and we briefly classify adaptive models into five categories:
\begin{itemize}
    \item \textbf{Block/Layer Adaptive}: This method selects a portion of blocks/layers to conduct DNN inference as shown in Fig \ref{fig:adaptive}; 
    \item \textbf{Channel Adaptive}: This method deploys a portion of channels of each layer to conduct DNN inference as shown in Fig \ref{fig:adaptive};
    \item \textbf{Early Exit}: This method uses the intermediate result from early layers to conduct prediction and skips the left layers in the network as shown in Fig \ref{fig:adaptive};
    \item \textbf{Multi-branch}: This method has diverse kernels for extracting features and uses a combination of kernels to conduct DNN inference as shown in Fig \ref{fig:adaptive};
    \item \textbf{Attention}: This method employs attention mechanism \cite{attention} to find the importantly spatial locations of images and only conduct computation-expensive convolution on these area as shown in Fig \ref{fig:attention}.
\end{itemize}

% Since conditional computation only activates a portion of the whole network, it allows to have a more complex model without significantly computational overhead. 

Runtime Network Pruning (RNP) \cite{rnp}, a channel adaptive method, is conceptually similar to \spt discussed in Section \ref{compress:pruning}. 
However, different from the methods in Section \ref{compress:pruning} which aim to produce a \textit{static} model at design time, RNP is a \textit{runtime and dynamic} pruning method, which uses RL to learn a policy to determine which filters should be used according to the input data. 
Feature Boosting and Suppression (FBS) \cite{gao2018dynamic} exploits channel saliency to judge which channels can be skipped, where a low overhead predictor is presented to predict channel saliency according to feature maps of the previous layer. 
Bejnordi \etal \cite{ehteshami2019batch} propose a channel-gate module to partially activate channels according to input. 
SlimmableNet \cite{slimmable,yu2019universally}, also a channel adaptive method,  proposes a method to train a network with different channel width configurations and the network can actively vary its channel configuration on the fly. 

Some works exploit attention mechanism to improve computational efficiency. Attention technique mimics human visual systems to locate their focal point on the important area and up-weights the areas of interest to improve the model accuracy. Attention has shown its successful application in CV \cite{attention} and NLP \cite{attentionNLP}. From efficient perspective, instead of up-weighting the important areas, some works avoid the computation-expensive convolution on the less important areas to improve the efficiency of DNN inference. 
SBnet \cite{sbnet} divides image into blocks with fixed size, and only convolves with blocks of interest. 
SACT \cite{sact} exploits adaptive computation time (ACT) technique \cite{act} to vary the computation per spatial location of input images. 
Verelst and Tuytelaars \cite{verelst2020dynamic} use Gumble-softmax \cite{gumbelsoftmax} to train a pixel-wise gate to identify the region of interest while Chen \etal \cite{chen2020dynamic} propose to combine attention with multi-branch technique to improve the capacity of lightweight models.  
CGNets \cite{cgnets} is conceptually similar to attention mechanism, where few feature maps
are used to identify the regions of interest within intermediate activations and the convolutional computation is only applied to the regions of interest. CGNets also provide a hardware implementation for accelerating its design.

Some works observe that for simple input data, the model can use intermediate results from early layers to predict the class without affecting the accuracy. This is called \textit{early-exit}. Adaptive Neural Network \cite{adaptiveNN} learns a policy to determine whether given an input data the model can use intermediate results from early convolutional layers to make accurate prediction. Based on this strategy, they also extend their policy to select a model from a set of models with different accuracy/latency trade-off. 
BranchyNet \cite{branchynet} uses \textit{early-exit} to skip some layers for easy input images, where a threshold is given to evaluate the confidence of predictions from early-exit.

Another common method to achieve adaptivity is \textit{multi-branch} which is conceptually analogous to mixture of experts \cite{jacobs1991adaptive}\cite{yuksel2012twenty}, where each branch has some convolution kernels and the multi-branch network selects some branches/convolution kernels per input to conduct inference. 
HydraNet \cite{hydranets} presents a multi-branch network which can select the best $k$ branches to conduct inference on per-input basis. 
Condconv \cite{condconv} proposes a conditional convolutional layer which combines different convolution kernels based on the input. 
Dynamic deep neural networks (D$^2$NN)  \cite{D2nn}, a bit different from \cite{hydranets,condconv}, formalize a network as a directly acyclic graph (DAG) which has different operations for each node within DAG and the network selects an effective and efficient execution path from DAG according to its input. D$^2$NN is trained in an end-to-end manner with assistant of RL.

ResNet \cite{he2016deep} is found to be tolerant to the removal of some blocks or layers without affecting the predictive accuracy \cite{veit2016residual}. Therefore, some works investigate how to bypass blocks/layers of a ResNet-similar network. SkipNet \cite{wang2018skipnet} uses previous layer's activation to determine whether the subsequent layer is required for the inference, where RNN and RL are used to control the executed blocks . Similarly, BlockDrop \cite{blockdrop} uses RL to train a policy network to determine the block configurations used for different input images. 
ConvNet-AIG \cite{convnet-aig} learns a gate module for each block of ResNets to select the inference blocks conditioned on input images. 

\begin{table}[]
    \centering
    \caption{Comparison of different adaptive DNN models}
    \resizebox{0.93\columnwidth}{!}{%
\begin{tabular}{c|c |c |c}
\hline
    Methods & Adaptive Scope & Training & Year   \\ \hline
     RNP\cite{rnp}&  Channels & RL  & 2017\\
     Benjordi \etal \cite{ehteshami2019batch}&  Channels & Gradient  & 2019\\
     SlimmableNet\cite{slimmable} & Channels & Gradient & 2019 \\
     FBS\cite{gao2018dynamic} & Channels & Gradient & 2018\\
     CGNets \cite{cgnets} & Channels & Gradient & 2019\\
     AdaptiveNN\cite{adaptiveNN} & Early-exit & Gradient & 2017 \\
     BranchyNet\cite{branchynet} &  Early-exit & Gradient & 2017\\
     D$^2$NN \cite{D2nn} & Multi-branch & RL & 2018\\
     Condconv\cite{condconv} & Multi-branch & Gradient & 2019\\
     HydraNets\cite{hydranets} & Multi-branch & Gradient & 2018\\
     SkipNet\cite{wang2018skipnet} &  Layers & RNN/RL & 2018\\
     BlockDrop\cite{blockdrop} &  Layers & RL & 2018\\
     SBnet \cite{sbnet} & Attention & Gradient  &2018\\
     Verelst and Tuytelaars \cite{verelst2020dynamic} & Attention & Gradient & 2020 \\
     Chen \etal \cite{chen2020dynamic} & Attention & Gradient & 2020 \\

    \hline
\end{tabular}
    }
    \label{tab:adaptive}
\end{table}

\noindent\textbf{Discussion:} Table \ref{tab:adaptive} summarizes the works discussed in this section. 
Adaptive models are attracting more attention from researchers due to the computation limit of resource constrained systems and highly dynamic environments. 
% Adaptive models are really useful for such dynamic environments. 
The advantages of adaptive models are twofold: 1) Adaptive models provide a flexible way to achieve trade-off between accuracy and efficiency on the fly; 
2) for some lightweight models adaptive models can increase the capacity of models, thereby improving the accuracy of the model without increasing the computational cost (because it only needs to partially activates the network). 
The existing adaptive models mainly consider the dynamics of input images, i.e., fewer channels or layers for `\textit{easy}' images and more channels or layers for `\textit{complex}' images. 
However, on edge systems, some system dynamics, memory contention, cache miss, bandwidth unavailability, etc,  also impact the execution of DNN application and this part is ignored in current research outcomes. \textit{For edge intelligence systems, it is of importance to take into account both dynamics, generating an adaptive edge intelligence system which can guarantee a certain level of QoS or meet real-time requirement under dynamic environments. }

\section{Discussion and Envisions}
\label{section:discussion}
DNN-based AI applications are increasingly integrated into our life and work and will greatly revolutionize the way we live and work. 
Some AI applications rely on super computing power to complete complex tasks while others will operate in proximity to data and end-users to help us live `smarter' and work `intelligently'. Edge intelligence, deemed as one of the most important AI trends, will make the proximity AI possible and accessible. In the first DNN decade (2011-2020), researchers around the world have designed many compelling DNN models,  
applicable to various domains, like nature language processing \cite{nlp}, AI-assisted medicine \cite{aimed}, robots, etc.
As reported in \cite{Intel}, over the years to come, the training/inference ratio of DNN models will increase to 1:5 from current 1:1 and enterprises will gradually add AI services to their core business so that they can profit from AI research and in turns AI applications can benefit the whole society. We can envision that the next important development for DNNs will be the practical deployment, especially like edge devices. To achieve this, we may need new DNN design methods, novel hardware \cite{dally2020domain} and the seamless cooperation between software and hardware. 

% Nevertheless, we can identify some techniques which may revolutionize our daily life,  
% The critical importance of DNN systems, or more general ML models on real systems has garnered increasing attention from both ML community and systems community \cite{mlsys}. How  In addition, edge intelligence is still in a golden 

% we can envision the significant potential of edge intelligence. 

\begin{figure}
    \centering
    \includegraphics[width=0.8\columnwidth]{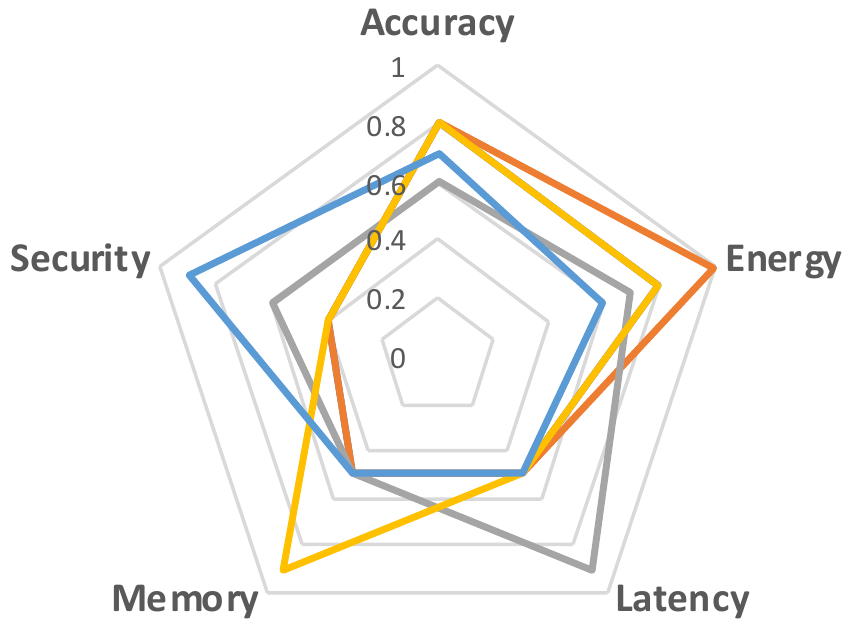}
    \caption{The real design space for Edge intelligent systems. There are various scenarios which have different design concerns.}
    \label{fig:design_space}
\end{figure}

In this section, based on our observations,  we identify some important topics which we believe will play a pivotal role in pushing DNNs to edge. 

\subsection{Different metrics oriented DNN models}
The accuracy improvement has been the highest priority in DNN research, where different models are justified by their accuracy increase, likewise a couple of percentiles. However when they are deployed on real systems, more metrics such as latency and power also matter. 
Moreover, DNN models are found to be vulnerable to adversary attack \cite{yuan2019adversarial}, so security is emerging as another new design concern for ML systems. 
Therefore, when designing DNN models for the emerging edge intelligent systems, the objective should be focused not only on accuracy but also on other critical metrics to have an overall consideration. 
More specifically, edge DNN models are really application-dependent or context-dependent, and it needs to find a good trade-off within the multi-dimension design space as shown in Fig. \ref{fig:design_space}.
As seen in Fig. \ref{fig:overview}, edge computing spans a wide spectrum. Some scenarios like UAVs, self-driving cars, robots, etc, have restrict requirements for accuracy, latency, and security, hence we may need a design point which is able to strike the balance within the design space.
To do so, we need to correctly use metrics or define new combination metrics for edge DNN systems. 

\begin{figure}
    \centering
    \includegraphics[width=0.8\columnwidth]{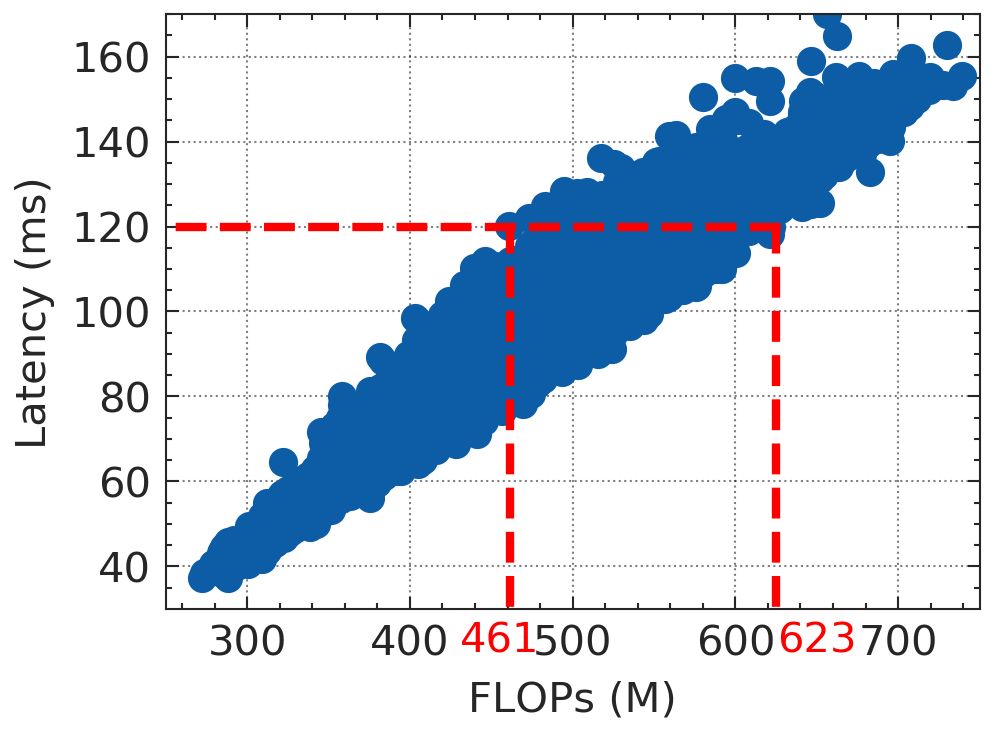}
    \caption{Latency vs FLOPs: The latency measurements of 10000 different architectures with different FLOPs count. The experimental platform is Nvidia Jetson Xavier, a widely used edge platform for robots and autonomous driving.}
    \label{fig:xavier_flops}
\end{figure}

For example, DNN research mainly adopts FLOPs as the indicator of model complexity, while FLOPs are used as constraints for network design and compression. 
However, the number of FLOPs may not directly translate to its latency, because DNN models have diverse architectures which may demonstrate different effectiveness on different hardware platforms. 
Fig. \ref{fig:xavier_flops} shows the latency of 100000 network architectures generated using DARTS \cite{liu2018darts} which are measured on Nvidia Jetson Xavier \cite{nvidia_jetson}, where we can see that the models with the same latency may differ in the number of FLOPs by up to 26\% (from 461 FLOPs to 623 FLOPs with latency 120ms) and the models with the same FLOPs perform different latency ranging from below 80ms up to 120ms. Therefore, we should carefully use the indirect metric to guide edge DNN design. 
In addition, design space for DNN is too large to have an exhaustive search. More design concerns will exacerbate this design complex issue. However, we still lack the measurement of the trade-off between each metric and this will lead to either high searching cost or sub-optimal design result. Thus, it will be necessary to define some combined metrics which can quantitatively evaluate the trade-off between different metrics, like energy-delay-product for conventional applications on CPU.

% nowadays, from our experimental results
% Many works on designing light-weight DNNs 

\begin{figure*}
    \centering
    \includegraphics[width=0.90\linewidth]{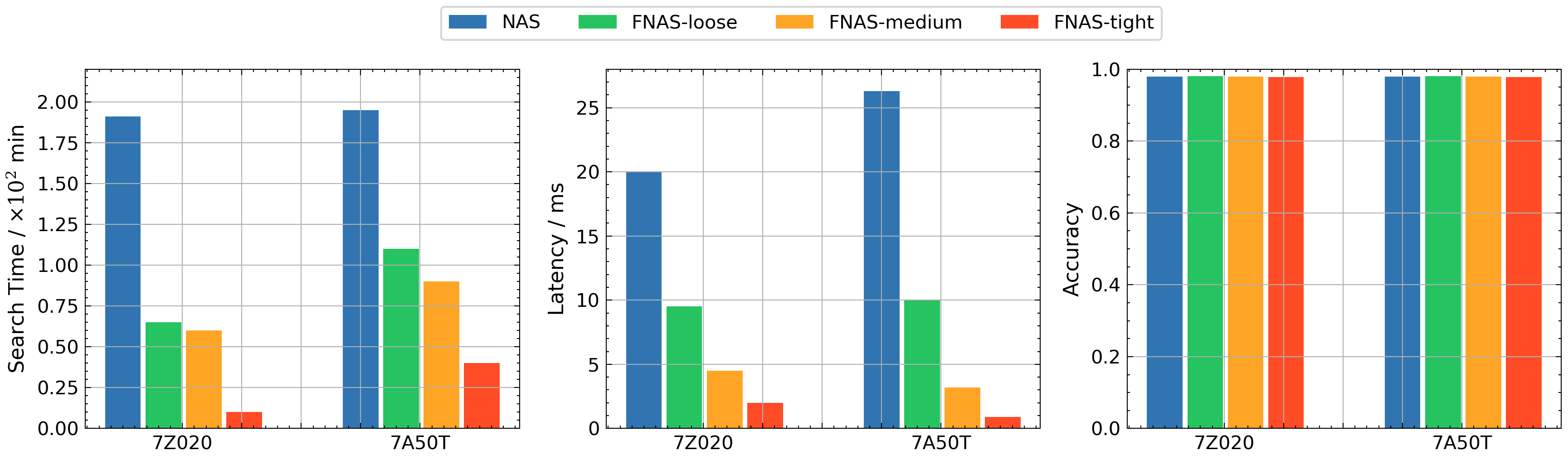}
    \caption{Effectiveness of Hardware-Software Co-Design Paradigm as shown in FNAS \cite{jiang2019accuracy}. FNAS-loose, FNAS-medium, and FNAS-tight denote three different design patterns in terms of the resource constraint.}
    \label{fig:hscodesign}
\end{figure*}

\subsection{Hardware-software co-design}
The high complexity of DNN models has spurred hardware architecture innovation to boost DNN training and inference over the past five years. The academia and industry have a consensus that the breakthroughs in hardware architecture will bring DNNs to a new level and boost DNN adoption \cite{tpu}\cite{hennessy2019new}. However, different hardware accelerators features diverse underlying characteristics, and the majority of DNNs were designed without consideration of underlying hardware features. We would like to call it \textit{hardware-agnostic design}.

\begin{figure}
    \centering
    \includegraphics[width=0.95\columnwidth]{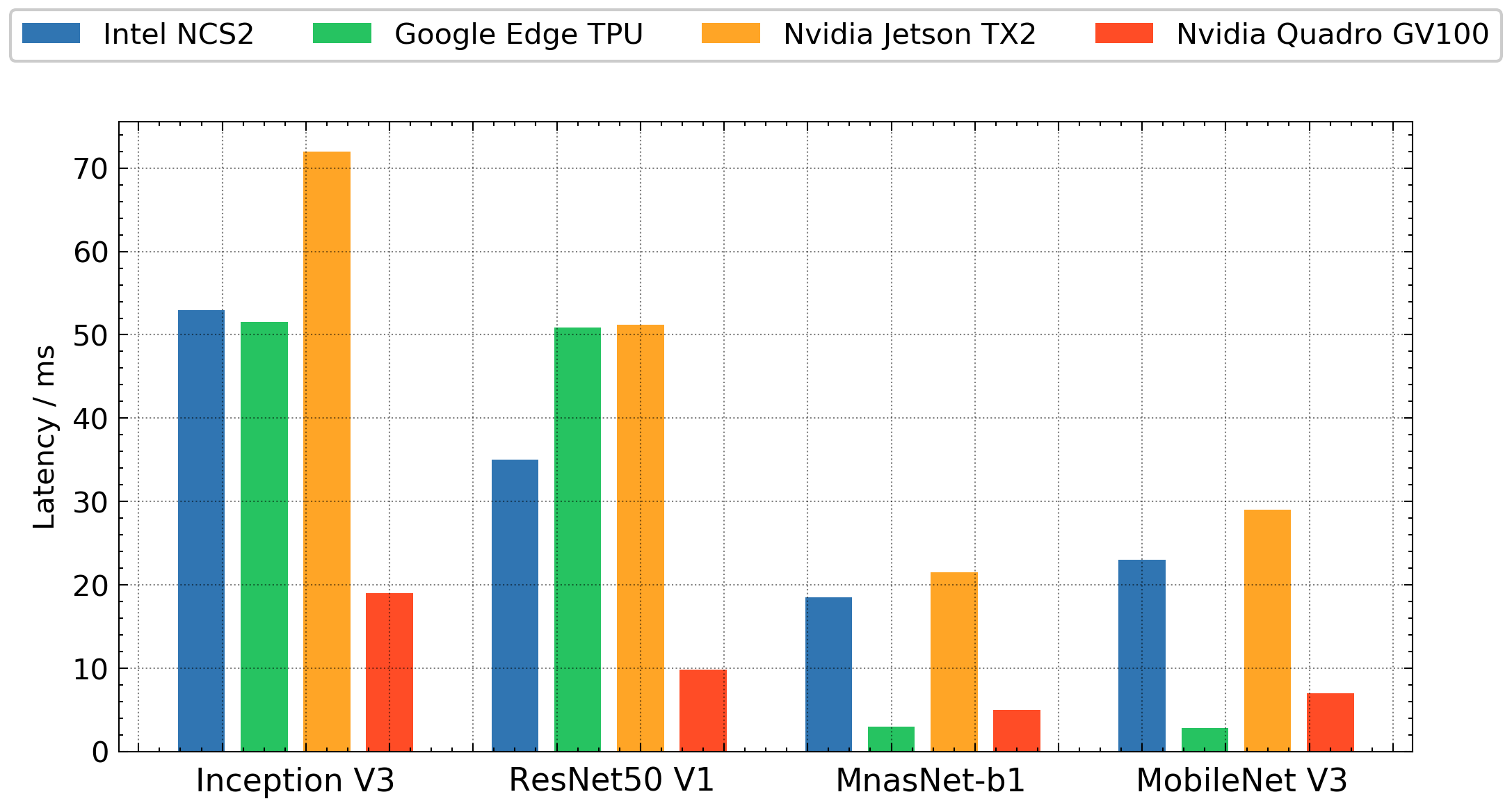}
    \caption{Latency of four state-of-the-art DNN models on four devices.}
    \label{fig:latency_4}
\end{figure}

It is known that accuracy of DNN models are highly related to its width (channels or feature maps) and depth (layers or blocks) \cite{tan2019efficient}, DNN models designed without hardware consideration may not fully utilize the underlying hardware. 
Fig \ref{fig:latency_4} shows the latency of four state-of-the-art DNN models, \textsf{Inception v3} \cite{inception}, \textsf{ResNet50} \cite{he2016deep}, \textsf{MnasNet} \cite{tan2019mnasnet} and \textsf{MobileNet V3} \cite{howard2019searching} on three edge devices, \textit{Intel Neural Computing Stick 2} \cite{intel_NCS2}, \textit{Nvidia Jetson TX2} \cite{nvidia_jetson}, \textit{Google Edge TPU} \cite{edgetpu} and one desktop GPU, \textit{Nvidia Quadro GV100}. 
For \textsf{MobileNet V3} on \textit{Edge TPU}, we use optimized models provided by Google, which has more MACs than the original models (990M vs 210M).     
From the experimental results, we see that \textsf{MnasNet} and \textsf{MobileNet V3} on \textit{Edge TPU} perform very low latency, even lower than the high-end GPU, because these two models are specifically optimized for \textit{Edge TPU}. 
These two models on \textit{Edge TPU} also consume higher power (5W) than \textsf{Inception V3} and \textsf{ResNet50} (4.6W). 
Based on the observations, we conjecture that since these two models are designed and optimized for \textit{Edge TPU}, they are able to better utilizes the parallelism of underlying hardware, thereby having lower latency and higher power consumption. In addition, as shown in \cite{aibenchmark}, the same neural architecture demonstrates up to 62x difference on modern mobile devices in terms of inference time. 

Some research strives to have a hardware-aware NAS, like FNAS \cite{jiang2019accuracy}, which directly incorporates resource constraints during implementing DNNs on FPGAs. The experimental results of FNAS are illustrated in Fig. \ref{fig:hscodesign}. 7Z020 and 7A50T denote the low-end and high-end FPGAs, respectively. FNAS can design models according to different resource constraints, i.e., FNAS-loose, FNAS-medium, and FNAS-tight. Thank to its hardware-aware method, FNAS can achieve the same accuracy level under different resource constraints, while significantly reducing the search cost and inference latency.

These together signal the importance of \textit{system-level hardware-software co-design}. The first decade of $21\text{th}$ century witnessed the emergence of system-level design methodologies \cite{gerstlauer2009electronic} for multicore system design. To alleviate the increasing complexity of multicore systems, different software and hardware co-design methods were proposed to elevate the design level to system-level by modeling software and hardware so that some tedious and error prone procedures can be avoided. The history may repeat for the emergent edge intelligent systems. \textit{The large and complex design space of edge intelligent systems need  hardware-software co-design to facilitate the effective and efficient design and implementation of edge intelligent systems.} 

To achieve co-design, we need to determine an effective design space for DNN models which will be helpful to reduce the costly design time.  We have seen some recent efforts towards defining the effective design space for DNN models \cite{radosavovic2019network,radosavovic2020designing}. 
At the same time, effective hardware modeling techniques are needed. We need metrics like 
 Roofline \cite{roofline} to guide the direction in finding efficient network upon a target platform and  
%  can be modified accordingly and be applied to evaluating the new types of hardware. 
a standard benchmark which can fairly and quantitatively evaluate various DNN models on new hardware accelerators, like MLPerf \cite{mlperf} and ParaDNN \cite{wang2020systematic}. 
In addition, to better utilize the underlying hardware, some DL compilers may need to be integrated into the co-design framework, like TVM \cite{chen2018tvm} and patDNN \cite{niu2020pat}.

\subsection{Lightweight models for other applications}
Currently, the majority of works regarding deep learning on edge systems target computer vision tasks, i.e., \textit{image classification} and \textit{object detection}.
% Throughout the paper, all literature we review focus on . We think there are two reasons for this. 
We have seen some successful CV-based edge intelligent systems, such as face recognition, object tracing on UAVs, navigation on robots, video analytic systems \cite{hussain2019intelligent}, etc. However, we also see the success of DNN models in other domain, like NLP, machine translation, etc. These models, like BERT \cite{devlin2018bert}, are known to be highly complex, even more complicated than DNN models for image classification, and a recent study in \cite{strubell2019energy} raises a concern regarding the environment effect of training complex NLP models. As diverse applications will be increasingly implemented on edge systems, we need new methods or frameworks to design light-weight DNN models for domains other than computer vision. Like \cite{liu2019point}, an efficient point-voxel CNN is proposed for efficient 3D learning, and this model can help to implement 3D AR/VR applications and SLAM \cite{durrant2006simultaneous} of autonomous driving on edge systems. Li \etal \cite{li2020gan} recently present a method to compress generative adversarial nets (GAN) \cite{goodfellow2014generative}. By means of compressed GAN, some GAN-based applications, e.g., style transfer, image synthesis, etc, can be efficiently implemented on edge systems. 
Some recent works study to compress the complex NLP models such that they can be deployed on resource-constrained edge systems \cite{litetransformer}\cite{tinybert}\cite{mobilebert}. 
Only few efforts are made towards designing lightweight DNN models of other domains for edge systems, but there is a huge potential to exploit such models on edge systems.

\subsection{Learning on The Edge}

In this paper, we review many techniques aiming to design lightweight models for edge intelligence systems, where the models are assumed to be trained on powerful servers but are deployed on edge devices for inference. 
However, due to high communication overhead, on-time model update and possible leakage of confidential data, some edge systems prefer to train the model locally, i.e., \textit{on-device learning} \cite{ondevice}. 
On-device learning is a challenging task, because training is a more computation-intensive and memory-hungry procedure compared to inference, whereas edge systems are resource-constrained in many settings. 
% On-device learning enables edge systems to become intelligent gradually without the server's instrument 
Moreover, limited energy supply of some edge systems will make this issue even more difficult.
Although few efforts have been made towards efficient learning at the edge, such as \cite{e2train,wu2020enabling}, learning at the edge is still at its early stage. There still remain a lot of issues to be addressed in this topic. 
% On resource-constrained edge systems, limited energy and memory expose new challenges to on-device learning.  
Some new methods, software frameworks and underlying libraries are needed to facilitate the effective and efficient training on edge systems with consideration of limitation and constraints imposed by edge systems, e.g., \cite{tinytransfer}.

% \begin{figure}
%     \centering
%     \includegraphics[width=0.95\columnwidth]{figures/Latency.pdf}
%     \caption{Latency of four state-of-the-art DNN models on four devices.}
%     \label{fig:latency_4}
% \end{figure}

 \subsection{The data challenge at edge} 
%  There are two ways to learn at the edge: 1) Learning from scratch---The model are completely trained on the devices with the intact data set; 2) Learning incrementally. 

The success of DNN models heavily relies on high-quality and large-scale datasets, such as ImageNet, but for some edge applications, e.g., edge surveillance systems in wild life, defect detection in manufacturing process, etc, it is difficult or expensive to collect massive amount of data and label them to train a good model.  
% At the same time, edge systems are expected to be deployed in different contexts for diverse purpose, from UAV to intelligent manufacturer as shown in Fig. \ref{fig:overview}, but 
Thus, the majority of edge intelligence systems without large-scale dataset exploit \textit{transfer learning} \cite{transfer} to learn a competitive model, 
where a model trained with large-scale dataset is provided to extract features and then the classification layer (fully-connected layer) is further fine-tuned according to domain-specific dataset such that the model can be adapted for the new domain. 

However, since, during the long-term operation, edge systems are likely to collect data with different distributions from the original training data or data pertaining to a new class which is not included in the original training data, 
we may need to update the model on edge systems in order to provide better prediction performance or infer a new class. On one hand, edge systems can update models locally by using \textit{incremental learning} \cite{incremental}\cite{shin2019incremental}, where techniques in \cite{lifelong} are deployed to improve the accuracy for new data and infer unknown classes. 
On the other hand, a group of edge systems is able to help each other to improve models by using \textit{federated learning} \cite{federated}. 
% sharing some useful information with each other, i.e., if we can assemble all data from different systems, we can probably have a large-scale data. Unfortunately, 
As discussed several times in this paper, data privacy is one motivation of edge systems, so it may be impossible to collect data from different edge systems and share them with each other. \textit{Federated learning} (FL) \cite{federated} is proposed to attack this issue, where instead of sharing the data with a centralized server clients (e.g., an edge system) in FL only share the learned gradient with others which will not leak data. 
Federated learning is a promising solution to share the information among several models or data sources while still keeping the confidential of data. Thus, FL can be used as a powerful tool to connect edge intelligence systems to improve their intelligent ability. Recently, few works study to employ incremental learning \cite{shin2019incremental} and federated learning \cite{wang2019edge}\cite{wang2019adaptive} with edge systems, but the research in this context is still in its infancy. 
The breakthrough in this area will pave the way of ubiquitously adoption of edge intelligence systems in our life.

\section{Conclusion}
\label{section:conclusion}
The convergence of edge computing and artificial intelligence leads to the concept of edge intelligence. Edge intelligence is in its early stage and needs sustainable efforts. This paper mainly surveys DL techniques which will facilitate the efficient deployment of DNN models on edge systems, i.e., lightweight models, network compression, hardware-aware NAS and adaptive models.
We provide some of our thoughts about edge intelligence systems and hope this paper can help researchers from edge computing community to understand the state-of-the-art DL techniques and to explore new opportunities in edge intelligence era. As stated in \cite{computational_limit}, DL algorithms are approaching the computational limits of computing systems and this probably indicates that designing efficient DNN models will soon become a standard not only for edge systems but also all AI systems. 
Then, designing efficient DNN models will become a mainstream.

%\IEEEpubidadjcol

%\subsubsection{Subsubsection Heading Here}
%Subsubsection text here.

% or


% \appendices

% % you can choose not to have a title for an appendix
% % if you want by leaving the argument blank
% \section{}

% \section*{Acknowledgment}

\bibliographystyle{IEEEtran}
\bibliography{reference}
\end{document}